\documentclass[a4paper]{cas-sc}
\usepackage[authoryear,longnamesfirst,square]{natbib}

\def\tsc#1{\csdef{#1}{\textsc{\lowercase{#1}}\xspace}}
\tsc{WGM}
\tsc{QE}

\usepackage{color}
\definecolor{mybar}{rgb}{1.0, 0.4, 0.0}

\usepackage{natbib}
\setcitestyle{numbers,square}
\usepackage{enumitem}
\usepackage{url}
\usepackage{hyperref}
\hypersetup{
	colorlinks=false,
	linkcolor=cyan,
	filecolor=blue,      
	urlcolor=red,
	citecolor=green,
}
\usepackage{graphicx}
\usepackage{subfig}
\usepackage{caption}
\usepackage{svg}
\usepackage{verbatim}
\usepackage{amsmath}
\usepackage{bbding}
\usepackage{booktabs}
\usepackage{algorithmic}

\usepackage{threeparttable}

\usepackage{amssymb}
\usepackage{multirow}
\usepackage{colortbl}
\usepackage{tabu}
\usepackage{pifont}
\usepackage{colortbl}
\usepackage[dvipsnames]{xcolor}
\usepackage{enumitem}
\usepackage{ragged2e}
\usepackage{color}
\usepackage{float}
\usepackage{booktabs}
\usepackage{multirow}

\usepackage[commandnameprefix=always]{changes}
\definechangesauthor[name={Per cusse}, color=orange]{per}

\usepackage{indentfirst}
\usepackage{xcolor}
\usepackage[linesnumbered,ruled,vlined]{algorithm2e}
\usepackage{balance}

\begin{document}
\let\WriteBookmarks\relax
\def\floatpagepagefraction{1}
\def\textpagefraction{.001}

\title[mode = title]{Noisy Label Processing for Classification: A Survey}



\author[1]{Mengting Li}

\address[1]{Beijing University of Posts and Telecommunications, No.10 Xitucheng Road, Haidian District, Beijing, China}

\author[2]{Chuang Zhu}
\address[2]{Beijing University of Posts and Telecommunications, No.10 Xitucheng Road, Haidian District, Beijing, China}

\cormark[2] 

\cortext[2]{Corresponding author: Chuang Zhu} 
\begin{abstract}[S U M M A R Y]
In recent years, deep neural networks (DNNs) have gained remarkable achievement in computer vision tasks, and the success of DNNs often depends greatly on the richness of data. However, the acquisition process of data and high-quality ground truth requires a lot of manpower and money. In the long, tedious process of data annotation, annotators are prone to make mistakes, resulting in incorrect labels of images, i.e., noisy labels. The emergence of noisy labels is inevitable. Moreover, since research shows that DNNs can easily fit noisy labels, the existence of noisy labels will cause significant damage to the model training process. Therefore, it is crucial to combat noisy labels for computer vision tasks, especially for classification tasks. In this survey, we first comprehensively review the evolution of different deep learning approaches for noisy label combating in the image classification task. In addition, we also review different noise patterns that have been proposed to design robust algorithms. Furthermore, we explore the inner pattern of real-world label noise and propose an algorithm to generate a synthetic label noise pattern guided by real-world data. We test the algorithm on the well-known real-world dataset CIFAR-10N to form a new real-world data-guided synthetic benchmark and evaluate some typical noise-robust methods on the benchmark.
\end{abstract}

\begin{keywords}
Classification, Label Noise, Deep Learning, Computer Vision, Noise Pattern.
\end{keywords}

\begin{highlights}
    \item Review deep learning approaches for noisy label processing in image classification 
		
	\item Methods based on semi-supervised learning are superior in noisy label processing

    \item Review different noise patterns proposed to design robust algorithms

	\item Currently proposed noise patterns are either far-from-reality or uncontrollable

	\item A real-world data-guided synthetic noise pattern fosters robust algorithm testing
\end{highlights}

\maketitle


%

\section{Introduction}

In recent years, deep neural networks (DNNs) have gained remarkable achievement in computer vision tasks, such as image classification, object detection, and segmentation. The success of DNNs depends a good deal on the richness of data. Large datasets like ImageNet \cite{krizhevsky2012imagenet} play an integral role in the development of DNNs. However, numerous and correctly labeled data is not always available. For non-expert sources, such as Amazon’s Mechanical Turk crowd-sourcing \cite{yu2018learning} and online queries \cite{xie2019improving}, it is highly likely that unreliable labels will occur in the label acquisition procedure. Moreover, even for experienced domain experts \cite{lloyd2004observer}, annotation can be a complicated and time-consuming process(e.g., in medical applications), which may lead to incorrect labels. Such unreliable labels are called noisy labels. This problem has attracted more attention in the field of machine learning because noisy labels are common and inevitable in real-world scenarios.

Noisy labels are corrupted from ground truth labels and they have an adverse impact on the performance of DNNs. Recent evidence suggests that DNNs have such strong capacity that they can easily fit noisy labels during the model learning process, resulting in poor generalization performance \cite{zhang2021understanding}. Thus, learning robustly with noisy labels can be a challenging task. It is crucial to process noisy labels for computer vision tasks, especially for classification tasks.

Studies over the past decades have provided critical solutions on how to learn with noisy labels. Beyond conventional machine learning methods, deep learning approaches have gained significant attention in computer vision, such as designing robust architecture, robust regularization, robust loss functions, and sample selection. As many solutions rely on noise detection and removal algorithms, more and more combined methods have emerged and gained significant success. In particular, methods combined with semi-supervised learning and contrastive learning techniques achieved state-of-the-art performance. It is meaningful to summarize those methods for researchers interested in the field of noisy labels and there have been some literature review in the past.

Fr´enay et al. \cite{frenay2013classification} provided a comprehensive survey on different families of noise-robust classification algorithms to deal with label noise. Moreover, they also discussed different definitions and potential consequences of label noise, as well as existing datasets and data generation methods. Filipe et al. \cite{cordeiro2020survey} described the main deep learning approaches that deal with label noise and analyzed state-of-the-art results. Meanwhile, Davood et al. \cite{karimi2020deep} focused on studies that trained deep neural networks in medical applications in the presence of label noise. Experimental results with three medical datasets were presented in the survey. Recently, Hwanjun et al. \cite{song2022learning} provided a comprehensive survey on recent deep learning methods for overcoming label noise and presented a systematic methodological comparison of existing approaches using six popular properties for evaluation. However, even if there are several label noise surveys, they mainly focus on the method review and none cover complete taxonomies of label noise patterns, which are important for algorithm testing and noise detection. Our survey mainly focuses on image classification problems in the presence of label noise and different noise patterns that have been proposed to design robust algorithms. Furthermore, this survey is the first attempt to seek the inner pattern of real-world noise and propose a real-world data-guided type of synthetic label noise pattern. 

The major contributions of this survey are as follows:
\begin{enumerate}
    \item[a)] A systematic review of the evolution of different deep learning approaches for noisy label processing in the image classification task in the past five years. We review various approaches including noise transition matrix estimation, noise-robust regularization, sample selection, semi-supervised learning, and other hybrid methods.
    \item[b)] We review different noise patterns proposed to design robust algorithms and categorize them into three taxonomies: instance-independent, instance-dependent, and human real label noise.
    \item[c)] We suggest two indicators, i.e., noise transition matrix and feature concentration distribution, to represent the inner pattern of real-world label noise and propose an algorithm to generate a real-world data-guided synthetic label noise pattern.
    \item[d)] We test the algorithm on the well-known real-world dataset CIFAR-10N to form a new real-world data-guided synthetic benchmark. We evaluate the typical methods of each category and the recent state-of-the-art methods in the image classification task with label noise on the benchmark we proposed for further research.
\end{enumerate}







The rest of this survey is organized as follows. In Section \uppercase\expandafter{\romannumeral2}, we discuss the definitions of noisy labels, as well as some preliminaries. Different types of noise patterns are depicted in Section \uppercase\expandafter{\romannumeral3}. Then, we summarize the real-world benchmark datasets that the image classification task widely used in Section \uppercase\expandafter{\romannumeral4}. In Section \uppercase\expandafter{\romannumeral5}, we present several types of approaches to deal with label noise. In Section \uppercase\expandafter{\romannumeral6}, we suggest the potential indicators of the pattern of real-world noise, design a real-world data-guided synthetic label noise generation algorithm and propose a noisy benchmark as an example. Subsequently, we evaluate the performance of some representative methods on the proposed benchmark in Section \uppercase\expandafter{\romannumeral7}. Finally, section \uppercase\expandafter{\romannumeral8} concludes this paper.

\section{Definition and Preliminaries}
\subsection{Definition of Label Noise and Survey Scope}
Noise can occur in different tasks, appearing in an image-wise label as in a classification problem or a pixel-wise label as in a segmentation problem. This survey focuses on discussing label noise under the image classification tasks. A classification task aims to predict classes for new data with a model trained from known data. In this survey, we assume that every training sample is associated with one observed label, which might differ from its true label. Before being presented to the algorithm, labels might be corrupted in many processes, such as annotation errors or subjectivity bias. Those corrupted labels are called noisy labels. We refer to noisy samples as those whose labels differ from true ones. In addition to label noise, feature noise or attribute noise also degrades the performance of DNNs. Two types of feature noise have been actively researched in recent years. First, adversarial attacks are proven to be a severe threat to the development of deep learning applications. While the small perturbations in the input data, so-called adversarial examples, are perceptible, they can negatively affect the training model. In addition, missing data imputation also gained much attention. Lost records, uncollected information, and many other reasons may result in the missing values in input data. The imputation algorithms focus on estimating the missing values from the observed data. Although feature noise is closely related to robust training by affecting the values in input data features, it is usually less harmful compared with label noise. In this survey, we only focus on the noise in the label rather than the input data. Furthermore, it should be noted that in the literature review section of this paper, our focus is mainly on novel methods in recent years, and we will no longer review methods that have not been updated much in recent years, such as robust loss functions. For those methods, readers can refer to  \cite{song2022learning} if necessary.
\subsection{Problem Statement}
This paper considers a typical supervised learning task, a $c$-class classification problem. Assume that we are provided with a training dataset $D=\{(x_{i},y_{i})\}_{i=1}^N$, where $x_{i}\in X$ is the $i_{th}$ image and $y_{i}\in Y$ is its corresponding label in a one-hot manner. $X$ denotes the data feature space, and $Y$ denotes the label space over c classes. Specifically, $y$ refers to the image’s observed label. $y$ might be different from the true label because noise can appear in the annotation process in a noisy label scenario. We denote $y^*\in Y^* $as the true label over c classes. We denote the probability distribution of class $y$ for instance $x$ as $p(y|x)$, and $\sum_{i=1}^c p(y_{i}|x)=1$ . A classification task aims to learn a classifier $f$ that maps the data space to the label space, finding the optimal parameter to minimize the empirical risk $R$, which is defined by the loss function \cite{cordeiro2020survey}. 
\begin{equation}
    R= E_{(x,y) \in D}[L(f(x),y)]=E_{x,y_{x}}[L(f(x),y_{x})] ,
\end{equation}
where we denote the loss function as $L$, the map function as $f$, and the expectation as $E$.

As is shown above, the final obtained empirical risk is not noise-tolerant since the computing process of the loss function contains noisy labels. Thus, it is essential to design noise-robust algorithms to reduce the negative effect noisy labels have on the deep learning process.
\section{Noise Patterns}
This section presents three types of label noise that are designed for testing robust learning algorithms to model the real noisy scenario.
\begin{enumerate}
    \item[a)] \textbf{Instance-independent label noise:} This type of label noise modeling approach assumes that the label corruption process does not take account of the input data features. Symmetric noise and asymmetric noise are the representative models of this type. Symmetric noise flips the ground truth label to other classes with the same probabilities. In particular, symmetric noise is categorized into two groups: symm-inc noise and symm-exc noise \cite{kim2019nlnl}. We denote the noise rate as $\tau \in [0,1]$ and $\tau_{ij}$ as the probability that label $i$ is flipped to label $j$. Symm-inc noise is modeled by randomly choosing a label from all classes including the ground truth label, which means $\forall j \in A$, $\tau_{ij} = \tau/(c-1)$. In symm-exc noise, the ground truth label is not included in the label flipping options, where $i\neq j$, $\tau_{ij} = \tau/(c-1)$. In contrast to symmetric noise, asymmetric noise  \cite{patrini2017making}represents a noise process where the ground truth label is flipped into one specific label whose class is closer to the true label. For example, in the CIFAR10 dataset, the class \textit{bird} is more likely to be confused with a \textit{plane} instead of a \textit{cat}, as mapped by \cite{zhang2018generalized}. In MNIST,  $2 \to 7$, $3 \to 8$, $7 \to 1$, and $5 \to 6$ were mapped, following \cite{patrini2017making} . For asymmetric noise, $\tau_{ij}$ depends on the class of the instance. Although this type of noise generation approach can not model realistic label noise, it is still the baseline for learning algorithms with noisy labels due to the simplicity of the noise model and the easy generation process.
    \item[b)] \textbf{Instance-dependent label noise:} In order to model real noisy scenarios, instance-dependent label noise was proposed \cite{xia2020part} \cite{zhang2021learning} \cite{chen2021beyond}. This type of noise generation approach depends on both class label and input data features. Images that belong to the same class can be very different in the input features, affecting each instance’s chance of mislabeling. This phenomenon demonstrates that the flipping probability should highly depend on every specific instance. One early proposed instance-dependent label noise is called \emph{Part-dependent Label Noise} \cite{xia2020part}. Inspired by the theory that humans often recognize an instance based on observing the parts rather than the whole instance, \emph{Part-dependent Label Noise} utilizes anchor points to learn the transition matrices for the parts of the instance and subsequently aggregates all the parts transition matrices to infer the instance-dependent matrix for this instance. Thus, the inferred instance-dependent matrix can be exploited to generate instance-dependent noisy labels. Later on, Zhang et al. \cite{zhang2021learning} introduce a novel noise assumption called polynomial margin diminishing noise (\emph{PMD noise}), proposing a series of noise functions that allow arbitrary noise strength in a wide buffer near the decision boundary. Different types of noise functions are then utilized to make feature-dependent label noise corruption. Recently, an accessible but valid controllable instance-dependent noise synthesizing algorithm has been proposed in \cite{chen2021beyond}. The algorithm is designed by taking advantage of the prediction errors of DNNs. They normally train a DNN on a clean dataset for $T$ epochs and obtain different classifiers with various classification performances. The score of mislabeling and the potential noisy labels are computed by the average predictions of classifiers. The synthesized noise is instance-dependent through this process since the prediction errors are obtained from each instance. However, despite the Instance-dependent noise pattern taking images into consideration when choosing the flipping labels, the synthesizing process only simulates the deep neural networks making mistakes instead of humans. 
    \item[c)] \textbf{Real-world human annotated label noise:} Although the above synthetic label noise models are widely used as baselines for robust algorithm testing experiments, they can’t adequately model the real noisy scenario. Thus, real-world human annotated label noise is proposed by  \cite{wei2021learning}. Images from training datasets of CIFAR-10 and CIFAR-100 are re-annotated by human workers in Amazon Mechanical Turk. After a label selection process, the final obtained labels are called real-world human annotated label noise for CIFAR-10 and CIFAR-100. Thus, two new benchmarks are presented in \cite{wei2021learning}: CIFAR-10N and CIFAR-100N. Both datasets strongly reflect the real-world noisy scenario since the labels are carefully collected from Amazon Mechanical Turk. Moreover, while real-world noisy datasets, such as Clothing1M, suffer from problems of lacking clean label verification and large-scale complex tasks, CIFAR-10 and CIFAR-100 equipped with real-world human annotated label noise are more controllable and verifiable since there are ground truth labels for the whole datasets and various noise ratio can be obtained from different annotation selecting strategies. This type of noise helps establish controllable, verifiable and right-sized real-world datasets with both noisy labels and ground truth labels, which is beneficial for future research on learning with noisy labels. However, the acquisition of the real-world human annotated label noise for large-scale datasets is quite expensive and time-consuming.
\end{enumerate}

It can be seen that there are many challenges in constructing a suitable and close-to-reality noisy dataset in a quick and human-convenient way. To address this issue, we propose an algorithm to generate a real-world data-guided synthetic label noise pattern for dataset construction in Section VI.

\section{A Review on Real Noise Classification Datasets}
During the past few years, several publicly available datasets with real-world noisy samples have been proposed in different scenarios to help evaluate advanced methods for classification performance in this field. We will make a brief introduction to these datasets in this section.
\subsection{ANIMAL-10N}
The ANIMAL-10N \footnote{\href{https://dm.kaist.ac.kr/datasets/animal-10n}{https://dm.kaist.ac.kr/datasets/animal-10n}} dataset is composed of 55000 images for 10 kinds of animals. The 10 kinds of animals are divided into five sets as following five pairs of confusing animals:\{(cat, lynx), (jaguar, cheetah), (wolf, coyote), (chimpanzee, orangutan), (hamster, guinea pig)\}, where every two classes in one set are similar and hard to distinguish. The source of images is from online search engines. In particular, animal names are searched as the keyword in Bing and Google. 6000 images are crawled for each of the ten kinds to make up the final 60000 collected images. For the labeling process, 15 participants, including 10 undergraduate students and 5 graduate students, are recruited from KAIST online community to perform annotation. Every participant is asked to annotate 4000 images after one hour of training. Removing irrelevant images, about 55,000 images are reserved in the end, dividing 5,000 into the test set and 50,000 into the training set. Besides, images are almost evenly distributed in ten classes. The estimated noise ratio of ANIMAL-10N is about 8\%.
\subsection{Food-101N}
The Food-101N \footnote{\href{https://kuanghuei.github.io/Food-101N}{https://kuanghuei.github.io/Food-101N}} dataset comprises 310,009 images of 101 different types of food recipes, sharing the same categories as the Food-101 Dataset. The images were collected from Google, Bing, Yelp and TripAdvisor, with the keyword of Food-101 taxonomy. The website foodspotting.com is avoided in the collection process since it is where images of Food-101 come from. Each image is assigned a class label, which might be incorrect. Meanwhile, 60,000 verification labels, which represent whether the class label is correct or not in the form of 0 or 1, are manually added for a portion of training and validation samples for the targeted noise detection. The test set of Food-101 is used for the classification performance evaluation of Food-101N. The estimated noise ratio of Food-101N is about 18.4\%.
\subsection{Clothing1M}
The Clothing1M \footnote{\href{https://www.floydhub.com/lukasmyth/datasets/clothing1m}{https://www.floydhub.com/lukasmyth/datasets/clothing1m}} dataset is a clothing dataset of large scale in 14 classes: T-shirt, Shirt, Knitwear, Chiffon, Sweater, Hoodie, Windbreaker, Jacket, Down Coat, Suit, Shawl, Dress, Vest, and Underwear. All images are crawled from a few online shopping websites, and the surrounding texts are obtained as well for follow-up label selection. In particular, if the surrounding texts of an image contain only one keyword in the aforementioned 14 classes, this keyword will be chosen as its class label. Otherwise, the image will be discarded. Thus, all labels can be incorrect and noisy. Subsequently, a small portion of images are assigned manually refined clean labels and are divided into training, validation and test sets, containing 47,570, 14,313 and 10,526 images respectively. The remaining 1000,000 samples are treated as the training set. The estimated noise ratio of Clothing1M is about 38.5\%.





\subsection{WebVision}
The WebVision \footnote{\href{https://data.vision.ee.ethz.ch/cvl/webvision/download.html}{https://data.vision.ee.ethz.ch/cvl/webvision/download.html}} dataset consists of 2,539,574 images representing 1000 categories, the same as in the ILSVRC 2012 dataset. Images are queried on two popular online websites, the Google Image Search website and the Flickr website. To enhance algorithm development, 100,000 samples are split into a test set and a validation set, which respectively consists of 50,000 images. Samples of test set and validation set are sent to the Amazon Mechanical Turk (AMT) platform for human verification. The verification label for each sample, representing if the label is correct or not, is thus obtained. The estimated noise ratio of WebVision is about 20\%.
\section{Noisy Label Processing Category}

\begin{figure*}[!h]
    \centering
    \includegraphics[width=5.2in]{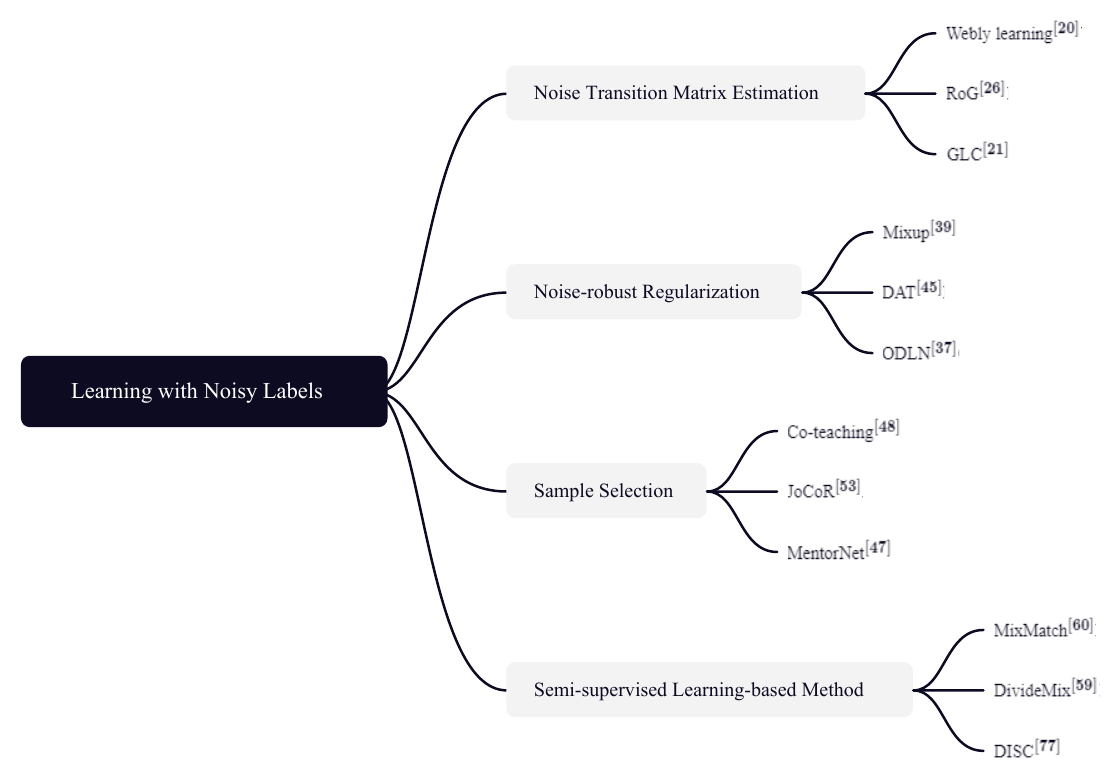}
    \captionsetup{justification=centering}
    \caption{A review of learning with noisy labels for classification. Specifically, we focus on four categories and present some of the typical algorithms.}
    \label{fig:overview}
\end{figure*}

\subsection{Noise Transition Matrix Estimation}
The first proposed methods for dealing with noisy labels are mostly noise transition matrix estimation, which learns the flipping probabilities between true and noisy labels, mimicking the noise process. Given a training dataset $D=\{(x_{n},y_{n})\}_{n=1}^N$, $y_{n} \in \{1,2,...,j,...\}$and a DNN with parameter $\theta$, the cross-entropy loss with noise transition matrix $T$ is expressed by
\begin{equation}
\begin{aligned}
    L(\theta)&=\frac1N \sum_{n=1}^N-log 
 p(y=y_{n}\mid x_{n},\theta) \\
    &=\frac1N \sum_{n=1}^N-log (\sum_{i}^c T_{ij} p(y^\ast=i\mid x_{n},\theta)) , \\
\end{aligned}
\end{equation}
where$$T_{ij}=p(y=j\mid y^\ast=i)$$

\begin{figure}[!h]
    \centering
    \includegraphics[width=5in]{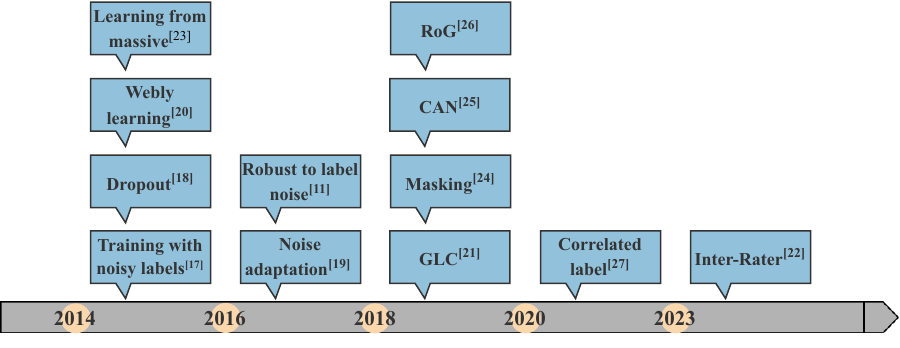}
    \captionsetup{justification=centering}
    \caption{The development of noise transition matrix estimation method.}
    \label{fig:overview}
\end{figure}

\underline{Literature review}: \emph{Training with noisy labels} \cite{sukhbaatar2014training} designs a learnable linear "noise" layer to represent the transition matrix and add it on top of the softmax layer to model the label corruption process. The method of \emph{Dropout} \cite{srivastava2014dropout} is similar to the one of \emph{Training with noisy labels} but adds the regularization skill of Dropout in the noise adaptation layer. Similarly, \emph{Noise adaptation layer} \cite{goldberger2016training} has the same basic ideas as \emph{Training with noisy labels} but adds another softmax layer on top of the noise layer and uses the pre-trained parameters for the initialization of the transition matrix. In contrast, \emph{Webly learning} \cite{chen2015webly} proposes a two-staged webly learning approach. It first trains the CNN model with easy images retrieved by  Google search engine. Subsequently, the obtained parameters of the first CNN initialize the second CNN model for hard images, and the confusion matrix of the relationship graph discovered from the first stage is used as the initialization of the transition matrix. Meanwhile, \emph{Robust to label noise} \cite{patrini2017making} estimate the transition matrix based on the output of the network trained on noisy labels and implement the loss correction with the transition matrix. Based on the method of \emph{Robust to label noise}, \emph{GLC} \cite{hendrycks2018using} further uses the clean fraction of training data to estimate the matrix. Moreover, \emph{Inter-Rater} \cite{bucarelli2023leveraging} propose a strategy called inter-annotator agreement to estimate the noise transition matrix of different annotators and utilize this strategy to learn from the label noise distribution.

To tackle more complicated and challenging label noise, more studies designed special network architecture to model the probabilistic distribution of complex label noise. \emph{Learning from Massive} \cite{xiao2015learning} proposes a probabilistic graphic model to model the transition between original labels and noisy labels. In particular, two networks are trained in the system, predicting the transition matrix and the noise type respectively. Afterward, \emph{Masking} \cite{han2018masking} designs a human-assisted scheme and induces the noise transition matrix from the label alignment with the human cognition of unreliable labels. \emph{CAN} \cite{yao2018deep} proposes two novel layers for the noise network, with one estimating the quality embedding variable to adjust the label transition probabilities and the other inferring the training posterior by combining the prior predictions and noisy labels. Meanwhile, \emph{RoG} \cite{lee2019robust} is a representative method for utilizing a noise transition matrix strategy. To obtain a more robust decision boundary, RoG creates a generative classifier from the hidden feature spaces of DNNs. The parameters of the proposed classifier are estimated by the minimum covariance determinant. Recently, \emph{Correlated Label}  \cite{collier2021correlated} introduces an innovative probabilistic approach to address the issue of input-dependent label noise in large-scale image classification datasets. By incorporating a multivariate Normal distributed latent variable on the final hidden layer of a neural network classifier, the method effectively models the aleatoric uncertainty caused by label noise. 

\underline{Remark}: The common drawback of this category is that it treats all the training samples equally without identifying the wrongly labeled samples, which results in a large estimation error for the noise transition matrix, especially in severe noise \cite{xia2022extended}. Meanwhile, this kind of method fails in the applications of real-world datasets due to the invalid prior assumption \cite{han2019deep}.

\subsection{Noise-robust Regularization}
For designing noise-robust regularization, \emph{Annotator confusion}  \cite{Tanno2020Learning} assumes the existence of multiple annotators and introduces a regularized EM-based approach to model the label transition probability. In \emph{ELR} \cite{2020Early}, a regularization term was proposed to prevent memorization of the false labels implicitly.

Some canonical regularization skills have been widely used in the computer vision field to prevent the training model from overfitting, such as dropout, weight decay, data augmentation, and batch normalization. \emph{REGSL} \cite{li2021improved} shows that those techniques can also improve the generalization performance in the presence of noisy labels. However, those skills alone failed in the robustness against label noise when noise is heavy. Therefore, more methods taking advantage of 
advanced regularization techniques are further proposed in both an explicit and an implicit way.

\underline{Literature review}: To avoid overfitting noisy labels, \emph{Bilevel learning} \cite{jenni2018deep} proposes a regularization method by solving a bilevel optimization problem based on cross-validation principles. This method differs from the conventional optimal method since the regularization constraint is also an optimal problem. It adjusts the weights of the stochastic gradient descent with both the gradients of validation mini-batches and training mini-batches. Thus, the updated model parameters will also perform excellently on the validation set. In contrast, \emph{Annotator confusion} \cite{Tanno2020Learning} presents an algorithm where the individual annotators are modeled by confusion matrixes and ground-truth label distribution can be learned, by using only noisy labels. A regularization term is added to the loss function so that the matrix can approximate convergence to the true annotator confusion matrix. Meanwhile, \emph{Pre-training} \cite{hendrycks2019using} demonstrates that although pre-training achieves similar performance to training from scratch in most traditional computer vision tasks, it is more beneficial to model robustness in the scenario of label corruption. Abundant representations learned by pre-training efficiently prevent the model parameters from overfitting noisy labels. \emph{PHuber} \cite{menon2019can} proposes a loss-based gradient clipping, which differs from the standard gradient clipping since it is robust to label noise. This variant of gradient clipping equals adjusting the underlying loss function. \emph{Robust early learning} \cite{xia2020robust} divides the model parameters into critical parameters and non-critical parameters, which respectively stand for partial parameters useful for fitting clean labels and the ones for fitting noisy labels. Subsequently, different regularization rules are applied for different types of parameters, thus mitigating the memorization effect of noisy labels. \emph{ODLN} \cite{wei2021open} found that open-set noisy labels may enhance the model robustness against label noise. It presents a regularization skill by inducing noises into open-set samples and leveraging them for training, which may benefit the model convergence to a stably flat minimum and improve model performance in the label noise setting. 

There are also some implicit regularization skills such as \emph{Adversarial learning} \cite{goodfellow2014explaining}, which demonstrates the phenomenon that the linear nature of adversarial examples, instead of nonlinearity, results in the neural networks' low generalization. Motivated by this, it improves the model tolerance of noise by harnessing adversarial training. Meanwhile, 
\emph{Mixup} \cite{zhang2017mixup} is a regularization skill utilizing the data augmentation principle, which improves the generalization performance of DNN and alleviates the memorization effect of noisy labels. It trains the neural network with virtual examples, which are constructed as the linear mixture of a pair of images and labels randomly selected from the training samples. The constructed augment training pair $(\Tilde{x}_{ij},\Tilde{y}_{ij})$ is defined as follows:
\begin{equation}
\begin{aligned}
    \Tilde{x} = \lambda x_i + (1-\lambda) x_j \\
    \Tilde{y} = \lambda y_i + (1-\lambda) y_j , 
\end{aligned}
\end{equation}
where $(x_i,y_i)$ and $(x_j,y_j)$ are two samples randomly drawn from the training set, and $\lambda\in [0,1]$.

What's more, \emph{Regularizing neural networks} \cite{pereyra2017regularizing} evaluates two forms of regularization including label smoothing and confidence penalty, and leads to the conclusion that these regularization skills show great improvement of model performance in supervised learning. Based on the research of \emph{Regularizing neural networks}, \emph{Label smoothing} \cite{lukasik2020does} explores whether the regularization skill of label smoothing is still practical in the presence of label noise. By mixing the one-hot label with uniform label vectors in the training process, this method reduces the overfitting of noisy labels and is beneficial for distilling a robust model from noisy samples. \emph{OLS} \cite{zhang2021delving} proposes a new label smoothing strategy, which is one of the effective regularization skills, to eliminate the overfitting problem of noisy labels in DNN. In particular, the soft labels are generated according to the prediction probability of the target label and non-target label. In contrast, \emph{LR} \cite{lienen2021label} argues that there might be a risk of introducing unexpected bias into the learning process by using the label smoothing technique. Thus, it proposes a new strategy called label relaxation, turning class labels into a set of upper probability distributions. What's more, \emph{ALASCA} \cite{ko2023gift} uses an adaptive label smoothing technique and auxiliary classifiers to obtain a robust feature extractor.

By aligning the feature distribution of clean and noisy data, \emph{DAT} \cite{qu2021dat} enforces more effective feature extraction. Additionally, a novel metric is generated to quantify differences between the distributions. \emph{DAT} serves as a regularization technique, effectively alleviating overfitting caused by noisy labels. \emph{Mitigating memorization} \cite{cheng2023mitigating} attempts to alleviate the memorization effect by using a representation regularizer to restrict the feature space of the decoupled DNN encoder without changing the model architecture. 

\begin{figure}[!h]
    \centering
    \includegraphics[width=5.5in]{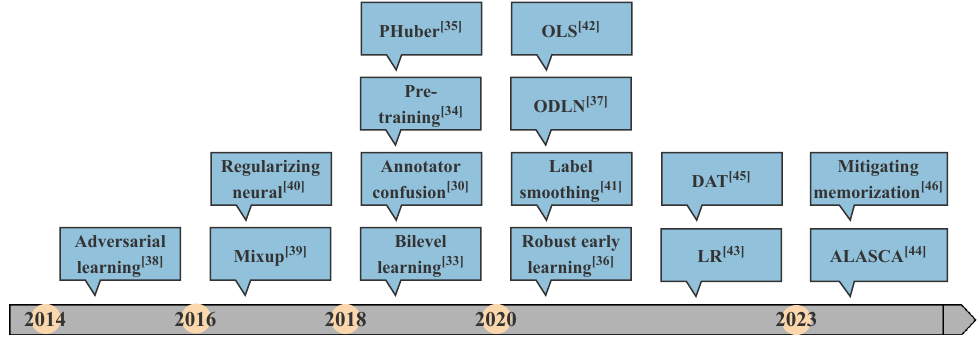}
    \captionsetup{justification=centering}
    \caption{The development of noise-robust regularization method.}
    \label{fig:overview}
\end{figure}

\underline{Remark}: This kind of method can easily be integrated with other noise-robust methods and training procedures because of its flexibility, requiring only simple modifications to the model architecture or training scheme. However, it suffers from either the problem of slow convergence or reduction of the model capacity. Some outstanding performances of the methods are conditionally limited or less obvious in other tasks such as image segmentation.

\subsection{Sample Selection}
Most of the recent successful sample selection strategies in the fourth category conducted noisy label processing by selecting clean samples through a "small-loss" strategy. \emph{MentorNet} \cite{jiang2018mentornet} pre-trained an extra network and then used the extra network for selecting clean instances to guide the training. The authors in \emph{Co-teaching} \cite{han2018co} proposed a Co-teaching scheme with two models, where each model selected a certain number of small-loss samples and fed them to its peer model for further training. Based on this scheme, \emph{INCV} \cite{chen2019understanding} tried to improve the performance by proposing an Iterative Noisy Cross-Validation (INCV) method. \emph{O2U-Net} \cite{huang2019o2u} adjusted the model's hyper-parameters to make its status transfer from overfitting to underfitting cyclically and recorded the history of sample training loss to select clean samples. This family of methods effectively avoids the risk of false correction by simply excluding unreliable samples. However, they may eliminate numerous useful samples.


To prevent the noisy samples from impairing the training performance, more and more advanced noise-robust methods adopt the strategy of sample selection. With this strategy, true-labeled samples are selected from the training set and leveraged for model updating. In the beginning, the memorization effect of DNNs is discovered, which suggests that DNNs tend to memorize the easy-patterned samples first and fit the noisy samples later. Based on this, the small-loss strategy is employed to select the clean samples whose training loss tends to be smaller in the beginning training epochs. However, with only a small-loss strategy, selection error may be accumulated, especially in the presence of ambiguous samples. Thus, on this basis, more methods propose multiple networks and multiple training rounds to avoid error accumulation and selection bias.

\underline{Literature review}: The main idea of  \emph{Decoupling} \cite{malach2017decoupling} is that it decouples the "when to update" principle from the "how to update" principle. In particular, it maintains two deep neural networks and selects samples based on the disagreement between the two networks. Besides, only the selected samples are updated. However, with the memorization effect of DNNs being discovered, more and more methods have started to adopt the so-called small loss trick to implement sample selection. Relative study suggests that noisy samples tend to exhibit larger loss in the early stage of training, when noisy samples are harder to memorize for DNNs. Based on this theory, the small loss trick treats samples with small loss as clean samples. \emph{MentorNet} \cite{jiang2018mentornet} advances curriculum learning by proposing another neural network, called MentorNet, to guide the training process of the original network, called StudentNet. Utilizing the small loss trick, the pre-trained Mentornet helps StudentNet to select possible true-labeled samples. In contrast, although \emph{Co-teaching} \cite{han2018co} trains two deep neural networks simultaneously as well, it makes them teach each other equally in the training process. In particular, in every mini-batch, both networks are fed full data and select possible clean samples based on the small loss trick for the other network. Then, two networks communicate with each other, and each chooses the samples that are needed for training and updating itself. \emph{Co-teaching+} \cite{yu2019does} achieves a further improvement by combing the \emph{Co-teaching} method with the disagreement trick in \emph{Decoupling}. Unlike \emph{Decoupling} and \emph{Co-teaching}, \emph{JoCoR} \cite{wei2020combating} doesn't adopt the disagreement trick and alleviates the problem of noisy labels from a different perspective. \emph{JoCoR} reduces the diversity of two networks by applying the co-regularization skill. Joint loss is proposed for every sample in the training procedure. Then the selected small-loss samples are used for updating both the two networks at the same time.

\emph{ITLM} \cite{shen2019learning} proposes a multi-round method to minimize the trimmed loss in each iteration based on the observation that clean and noisy samples have different training performances in the evolution of accuracy. Specifically, in the training scheme, selecting small-loss samples and using only those samples to retrain the model are alternatively implemented. Meanwhile, in \emph{INCV} \cite{chen2019understanding}, cross-validation is applied to randomly divide the noisy dataset in each training epoch, distinguishing most of the true-labeled samples. Subsequently, it adopts the Co-teaching method to train the model with identified clean samples in the latter training round. \emph{O2U-Net} \cite{huang2019o2u} records the loss of each sample in each iteration to determine the probability of a sample being noisy. Only some hyper-parameters of the model are adjusted to keep the training status underfitting other than overfitting in the cyclical training procedure.

To address the problem of open-set noisy labels, \emph{iterative detection} \cite{wang2018iterative} proposes an iterative learning scheme to detect noisy samples, utilizing the local outlier factor algorithm. While deep discriminate features are also learned in the iterative training process, a Siamese network and a reweighting module are employed to alleviate the side effects of noisy samples. In \emph{MORPH} \cite{song2021robust}, a self-transitional method is proposed, which suggests that the learning phase should be switched at a certain point in the training process. In the first phase, all samples are fed to the network to select clean samples. After switching to the second phase, only the selected samples are used for updating the network parameters. The clean set will be improved throughout the whole training round. Instead of using the small-loss trick as mentioned in most previous work, \emph{TopoFilter} \cite{wu2020topological} classifies true-labeled samples by taking advantage of the abundant representational information of data in the feature space. The obtained clean samples are of high quality due to the application of high-order topological information of data. In contrast, \emph{NGC} \cite{wu2021ngc} designs a graph-based framework, which detects the noisy samples by combining the probability of the noisy classifier and the data's geometric structure.

\begin{figure}[!h]
    \centering
    \includegraphics[width=5in]{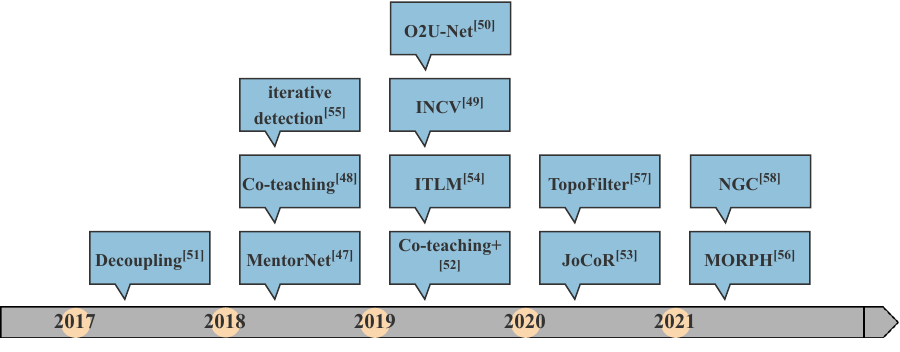}
    \captionsetup{justification=centering}
    \caption{The development of sample selection method.}
    \label{fig:overview}
\end{figure}

\underline{Remark}: In this kind of method, possible true-labeled samples can be discarded from the noisy dataset to benefit the training of DNNs and reduce the negative effect of noisy samples. However, selection errors might be accumulated through the sample selection process in some of the approaches. Moreover, only clean samples are used to update the model parameters. The abandonment of the remaining noisy samples results in the reduction of available data size, which is harmful to model training. Furthermore, the calculation cost and time cost will be increased in the multiple networks or multiple rounds kind of methods.

\subsection{Semi-supervised Learning-based Method}
To solve the shortcomings of the sample selection method, the methods of the fourth category based on semi-supervised learning treat the noisy samples as unlabeled samples and use the outputs of classification models as pseudo-labels for subsequent loss calculations. The authors in \cite{li2020dividemix} propose DivideMix, which relies on \emph{MixMatch} \cite{berthelot2019mixmatch} to combine training samples classified as clean or noisy linearly. \emph{LongReMix} \cite{cordeiro2023longremix} designs a two-stage method, which first finds the high-confidence samples and then uses the high-confidence samples to update the predicted clean set and train the model. Recently, \emph{Augmentation strategies} \cite{2021Augmentation} uses different data augmentation strategies in different steps to improve the performance of DivideMix, and \emph{C2D} \cite{2022contrast} uses a self-supervised pre-training method to improve the performance of DivideMix.

Specifically, for semi-supervised learning under label noise, the selected clean samples are treated as the labeled set, and noisy samples discard their labels to compose the unlabeled set. Two sets are then used to train DNNs in a semi-supervised way. For instance, in some methods, the model trained with clean samples will make predictions for unlabeled samples as pseudo labels. The unlabeled set with pseudo labels can then be treated as the labeled set to join the training process. The overview of the semi-supervised learning process (SSL) under label noise is shown in Figure 5.

\begin{figure}[!h]
    \centering
    \includegraphics[width=5in]{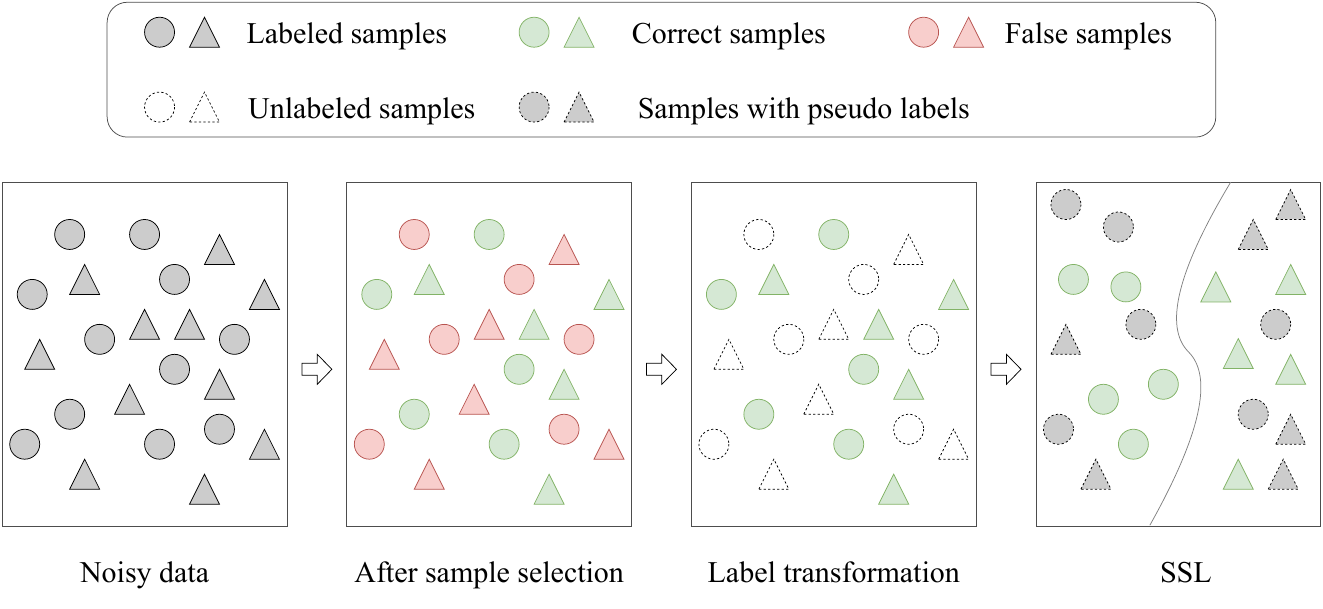}
    \caption{{The overview of the semi-supervised learning (SSL) process under label noise.
   }}
    \label{fig:SSL}
\end{figure}

\begin{figure*}[t]
    \centering
    \includegraphics[width=5.5in]{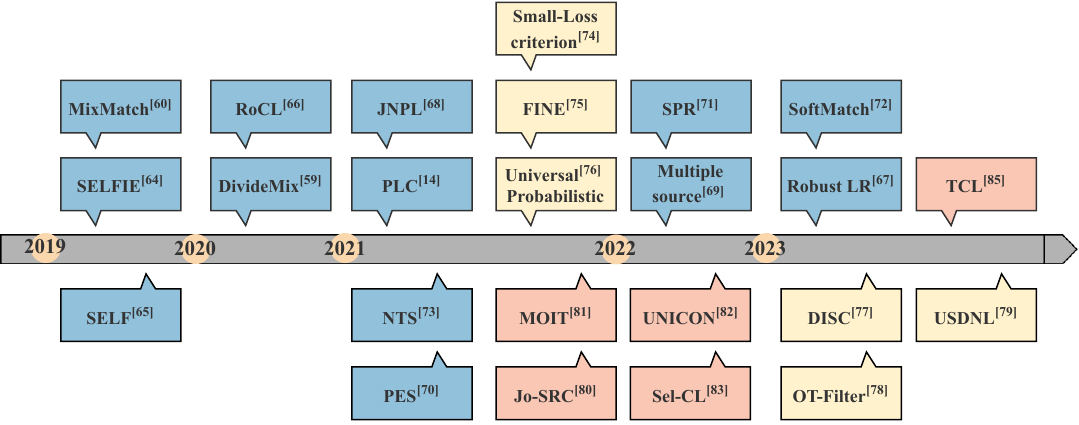}
    \captionsetup{justification=centering}
    \caption{The development of semi-supervised learning-based method. Blue, yellow and red rectangles represent typical semi-supervised methods, methods with novel techniques for sample selection, and methods combined with contrastive learning.}
    \label{fig:overview}
\end{figure*}

\underline{Literature review}: In \emph{SELFIE} \cite{song2019selfie}, after sample selection, the remaining false-labeled samples are refurbished by the selective loss correction strategy. Subsequently, the clean samples along with the adjusted samples will be trained together on the robust model. In contrast, \emph{MixMatch} \cite{berthelot2019mixmatch} combines different ideas of advanced semi-supervised learning methods to propose a new paradigm. Specifically, it predicts labels for unlabeled samples with the low-entropy trick and then leverages \emph{Mixup} to mix the labeled and unlabeled samples for further training. \emph{DivideMix} \cite{li2020dividemix} combines the noisy label learning framework with the semi-supervised learning. To be specific, it identifies the clean samples with a mixture model and then divides the noisy dataset into a labeled set and an unlabeled set, which is composed of clean samples and noisy samples respectively. Then, divided data will be fed to the network for training in a semi-supervised way with the help of \emph{MixMatch} strategy. Two networks are designed to be trained simultaneously to make the division for each other, avoiding the division bias. Meanwhile, \emph{SELF} \cite{nguyen2019self} filters the false-labeled samples by leveraging the average predictions of the whole training procedure instead of the latest epoch. Subsequently, the filtered samples will be trained with an unsupervised loss in a semi-supervised manner.
\emph{RoCL} \cite{zhou2020robust} updates the selection strategy to select samples based on both the loss and output consistency. It relabels the selected unlabeled samples by data augmentations and the dynamic training history. \emph{PLC} \cite{zhang2021learning} proposes a progressive algorithm to combat label noise by iteratively correcting noisy labels and refining the model. This algorithm can train a classifier to approximate the Bayes classifier. Discovering the drawback of confirmation bias in \emph{DivideMix}, \emph{Robust LR} \cite{chen2023two} proposes a technique combined with pseudo-labeling and confidence estimation to correct the noisy labels.

Besides, some novel views are suggested to combat the noisy labels. Inspired by using complementary labels, a new method called Joint Negative and Positive Learning (\emph{JNPL}) \cite{kim2021joint} is proposed, which unifies the filtering pipeline in a single stage and uses two loss functions in the process of semi-supervised training, NL+ and PL+, to train the CNN. The NL+ loss function enhances the convergence of noisy data, while the PL+ loss function enables faster convergence to clean data. \emph{Multiple source learning} \cite{silva2022noise} proposes a deep-learning framework to tackle noise from multiple sources, which is composed of two modules, MULTI-IDNC and MULTI-CCNC, to correct the instance-dependent and class-dependent label noise respectively. Moreover, \emph{PES} \cite{bai2021understanding} finds the phenomenon that different layers of DNN have different robustness against label noise. Thus, \emph{PES} proposes a framework to split the DNN into different parts, learning the former parts first and progressively training the latter parts in a semi-supervised manner to improve the overall performance of the model. \emph{SPR} \cite{wang2022scalable} proposes a penalized regression model to detect the noisy data, which are distinguished by the non-zero mean-shift parameters. As \emph{SPR} can be treated as a sample selection framework, the selected noisy data are further leveraged for semi-supervised learning. To resolve the quantity-quality trade-off issue in pseudo labels, \emph{SoftMatch} \cite{chen2023softmatch} designs a truncated Gaussian function to obtain a better version of the confidence threshold, thus generating superior pseudo labels. Besides, a uniform alignment strategy is proposed in the framework of \emph{SoftMatch} to solve the imbalance problem in pseudo labels. \emph{NTS} \cite{chen2021robustness} proved the robustness of the accuracy metric and the reliability of exploiting the noisy validation set. Based on those theories, it proposes a novel framework to select a teacher model according to the highest accuracy performance on the noisy validation set and further train a student model on model predictions of the teacher model.

Meanwhile, many techniques for sample selection in the semi-supervised learning process are proposed. \emph{Small-Loss criterion} \cite{gui2021towards} firstly discovers the theoretical reason why the widely used small-loss trick is effective in the field of learning from noisy labels. Furthermore, it enhances the small-loss principle based on the original version for better sample selection and combines it with the semi-supervised learning method \emph{MixMatch} to deal with noisy labels. Additionally, \emph{FINE} \cite{kim2021fine} leverages the eigendecomposition of the datagram matrix to evaluate the alignment between each data point’s representation distribution and its current dynamic representation. Then the obtained eigenvectors can be used for noisy data detection. \emph{FINE} can be collaborated with all kinds of noise-robust methods. A probabilistic model is proposed by \emph{Universal Probabilistic Model} \cite{wang2021tackling} to determine the confusing probability of each instance, thus identifying the confusing samples and correcting them with reliable labels learned by an alternating optimization algorithm. \emph{DISC} \cite{li2023disc} proposes a dynamic instance-specific strategy to dynamically calculate the selection threshold for each instance based on the model predictions in previous epochs. Then the noisy data can be divided into three subsets: clean, hard, and purified. Inspired by optimal transport, \emph{OT-Filter} \cite{feng2023ot} presents a sample selection technique using geometrical distances and distribution patterns to measure the discrepancy of training data. Furthermore, \emph{OT-Filter} can be collaborated with many current training schemes including semi-supervised learning. \emph{USDNL} \cite{xu2023usdnl} proposes to adopt cross-entropy-loss combined with the epistemic uncertainty, which is obtained through early training with single dropout, to select clean samples.

Recently, contrastive learning has gained much attention since it helps the training model obtain better feature representations. More studies have combined semi-supervised learning with contrastive learning to improve algorithm robustness. \emph{Jo-SRC}  \cite{yao2021jo}is introduced as a noise-robust approach that employs contrastive learning to train the network. Employing two different views of each sample helps the approach to determine whether it is clean or noisy and correct the noisy samples. Additionally, consistency regularization is introduced to enhance the performance of the model in terms of generalization. Meanwhile, \emph{MOIT} \cite{ortego2021multi} utilizes contrastive learning to obtain robust feature representations, thus generating more robust soft-labels. The disagreement between soft labels and original labels can then be exploited to distinguish the noisy samples for semi-supervised learning. \emph{UNICON} \cite{karim2022unicon} addresses the problem that clean samples are often selected from the easy classes and proposes a sample selection algorithm based on Jensen-Shannon divergence to eliminate the class-imbalance phenomenon in the selected set. Contrastive loss is also adopted in the training process to enhance unsupervised feature learning. \emph{Sel-CL} \cite{li2022selective} is the advanced version of \emph{Sup-CL} \cite{khosla2020supervised} (supervised contrast learning) for tackling the problem of noisy labels. Without the knowledge of noise rates, \emph{Sel-CL} measures the consistency between labels and representations to select the confident pairs for subsequent supervised contrastive learning. Moreover, \emph{TCL} \cite{huang2023twin} utilizes contrastive learning to obtain discriminative image features and proposes a Gaussian mixture model (GMM) over the features with the injection of model predictions. Out-of-distribution technique is also adopted to help with sample selection and correction.

\underline{Remark}: Semi-supervised learning-based methods fully exploit the training dataset. Therefore, test accuracy and model performance are greatly improved compared with approaches in the past. However, the complexity of these methods may introduce many more parameters and cost more computational resources.

\section{Noisy Data Generation Algorithm}
\begin{figure*}[!h]
    \centering
    \includegraphics[width=5.5in]{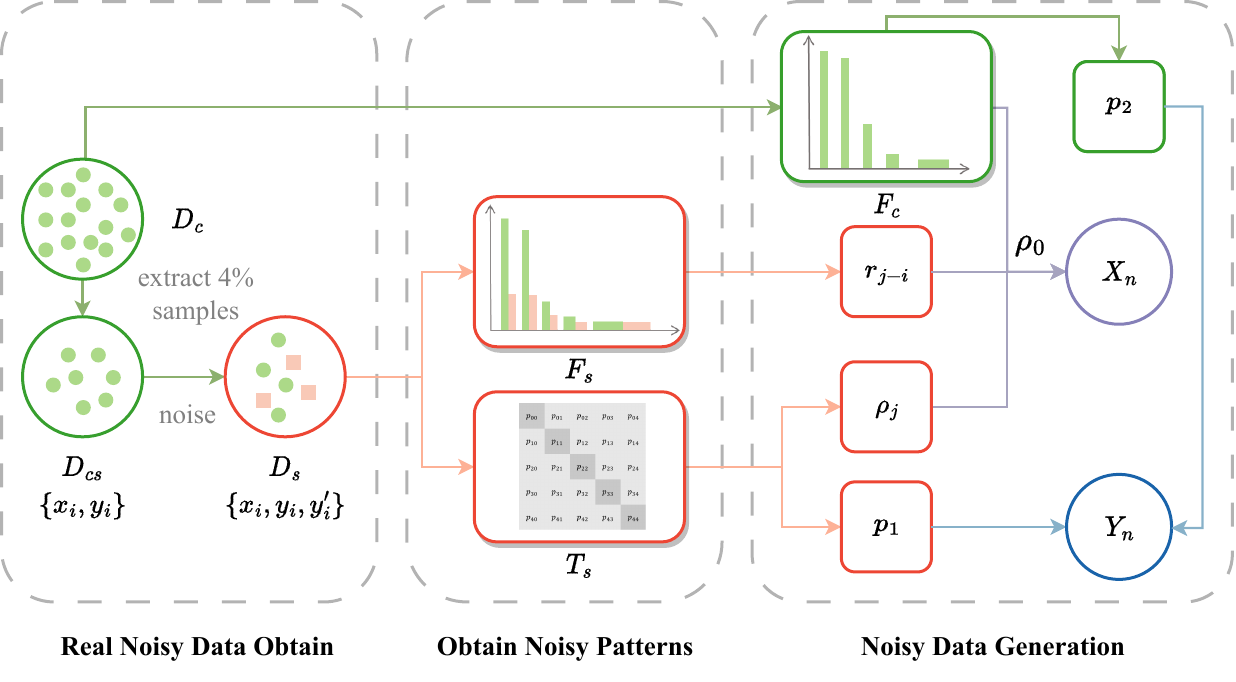}
    \caption{{The overview of RGN generation. Note that the green components represent the clean dataset and its corresponding feature outputs while the red ones represent the noisy subset and its its corresponding feature outputs. Purple and blue represent the generated noisy sample set and label set respectively. 4\% samples of the clean set $D_c$ are extracted to form the real noisy dataset $D_s$ by adding human annotations. Then the noisy patterns, i.e., the feature concentration distribution $F_s$ and the noise transition matrix $T_s$, are obtained from $D_s$. Meanwhile, feature concentration distribution $F_c$ of $D_c$ is also acquired, and the feature class probability $p_2$ is inferred. Subsequently, according to the noisy patterns, we can deduce the noise rate in each feature interval $r_{j-i}$, the noise ratio of each class $\rho_j$ and class label flipping probability $p_1$. In the end, the noisy samples $X_n$ are selected based on $F_c$, $r_{j-i}$, $\rho_j$ and $\rho_0$ (the expected overall noise ratio). The corresponding noisy labels of $X_n$ are selected based on $p_1$ and $p_2$.
   }}
    \label{fig:overview}
\end{figure*}
Since the emergence of the research of learning with noisy labels (LNL) , lots of efforts have been made to construct suitable and normative datasets with noisy labels for algorithm testing. The efforts are mainly reflected in two ways: real-world noisy datasets and synthetic noisy datasets. However, there are some drawbacks in both ways currently for constructing noisy datasets. 1) real-world noisy datasets: on the one hand, images and noisy labels are collected from the Internet or human annotation to construct the real-world noisy datasets. However, the building process of human-annotated datasets is extremely time-consuming and expensive while the website-crawling way can be quite unreliable and unstable. Besides, clean labels are unavailable in real-world noisy datasets, making it difficult to accomplish some important tasks for algorithm testing such as noise detection. Also, the noise ratio is fixed once the construction of the dataset is done, resulting in uncontrollable settings for flexible algorithm testing. Even though there is an attempt \cite{jiang2020beyond} to construct controllable real-world label noise by replacing the desired proportion of clean images with incorrectly labeled web images rather than changing the labels of clean samples, massive and expensive human annotation process for all images is still inevitable in this work. Moreover, it does not explore the pattern of real-world noise and lacks theoretical explanations. 2) synthetic noisy datasets: on the other hand, several noise injection procedures are adopted for clean image classification datasets to construct the synthetic noisy datasets, as mentioned in Section \uppercase\expandafter{\romannumeral3}. However, the symmetric and asymmetric noise pattern is far from fitting the real-world noise pattern due to the labels' independence of the images. Besides, despite the IDN noise pattern taking images into consideration when choosing the flipping labels, the synthesizing process only simulates the deep neural networks making mistakes instead of humans. 

It can be seen that there are many challenges in constructing a suitable and close-to-reality noisy dataset in a quick and human-convenient way. Besides, the number of currently available real-world noisy datasets is too few for the development of LNL. These limitations might severely hinder the data-driven algorithms from making more extensive progress because most of the algorithms are designed based on the theory of deep learning, which requires a large amount of data to train the model and avoid over-fitting.

Under this circumstance, we propose a new algorithm leveraging real-world information to inject controllable noise for clean datasets, generating real-data guided synthetic noise (RGN) with the characteristics of less time-consuming, easy-to-generate and close to real-world scenarios. This algorithm can be applied to any dataset in any noisy scenario. We focus on the literature review and noise generation algorithm design in this paper and build one such dataset based on CIFAR-10N as a reference to validate our algorithm. Researchers can further use the algorithm in other scenarios to generate different benchmarks. An overview of the algorithm is shown in Figure 7 and details are introduced in the following subsections.



\subsection{Real Noisy Data Obtain}

Firstly, around 4\% samples are extracted as a clean subset from the targeted clean set that needs the injection of label noise. Subsequently, we introduce the real-world label noise by annotating the clean subset by humans in a limited time. Thus, the real noisy data for the subset is obtained. We take CIFAR-10N, the human-annotated noisy version of CIFAR-10 dataset, for reference in the natural scenario. As both clean and noisy labels are known in CIFAR-10N, the human annotation process can be ignored. We denote the clean CIFAR-10 dataset as the clean set $D_{c}$, the CIFAR-10N dataset as the noisy set $D_{n}$, and 2000 images along with their noisy labels are sampled from the noisy set $D_{n}$ as the subset $D_{s}$. Specifically, 200 samples for each category are randomly selected from the shuffled noisy set. 

\subsection{Obtain Noisy Patterns}
Although great progress has been made in the field of noisy dataset generation, the real-world noise pattern still needs to be discovered. In this section, we propose a way to define the real-world noise pattern. Specifically, we first make some assumptions about noisy patterns and feature distributions based on past observations about label noise. To validate the assumptions, some mathematical models are proposed to define relative concepts about the feature distribution. Then we make a statistical analysis on the mathematical models of CIFAR-10N. In the end, we define the noise patterns based on the assumptions and validations.

\subsubsection{Observations and Assumptions}
 We first observe that the noise transition matrix has been widely leveraged for LNL algorithm designing. The noise transition matrix represents the probability of clean labels flipping to noisy labels for different classes. The matrix can be expressed as $T_{i,j} = \{Pr(\overline{y}=j|y=i)\} $. Abundant information such as overall noise ratio, noise ratio of each category and the flipping probability between two categories can be acquired from the matrix. Thus, we first introduce the noise transition matrix $T_{s}$ as one of the indicators of the real-world noise pattern.

In addition to the ground truth, the possibility of one image being misidentified as another is highly correlated with the feature of this image. Besides, how one image's feature pattern will affect its possibility of being mislabeled is still a problem to be discussed. Moreover, it is another question to decide which label its ground truth should be flipped to. To address these issues, we make the assumption that in the real-world scenario, the possibility of one image being mislabeled is related to the concentration degree of the image's features and we introduce the feature distribution $F_{s}$ as the other indicator of the real-world noise pattern. To be specific, if the feature is closer to the feature group of its ground truth and far away from the feature groups of other classes, this instance is less likely to be treated as a noisy sample. To validate this assumption, we implemented some experiments.

\subsubsection{Mathematical Models}
Before implementing the experiments, we first propose some mathematical models to define some relative concepts about the feature distribution $F_{s}$. For a specific image $I_{k}$, we define its total distance to other images with its class $C$ as $L_{intra}(I_{k})$, representing the degree of feature concentration within its class group, and the distance is defined as 
\begin{equation}
    L_{intra}(I_{k}) = \sum_{i = 1;i \neq k}^m\Vert I_{k} - I_{i}\Vert_{2}^2 , 
\end{equation}
where $m$ is the number of images in class $C$.

Meanwhile, the concentration degree of an image $I_{k}$ is also associated with its interference from other class groups. The interference can be expressed by $L_{inter}(I_{k})$, its total distance to images with any class label that is not $C$. $L_{inter}(I_{k})$ is defined as
\begin{equation} 
    L_{inter}(I_{k}) = \sum_{j = 1;j \neq C}^N\sum_{i = 1; y_{i} = j}^m\Vert I_{k} - I_{i}\Vert_{2}^2 , 
\end{equation}
where $N$ is the number of class labels.

Based on (4) and (5), the overall indicator $Con_{k}$ to analyze the feature concentration degree for image $I_{k}$ is defined as follows
\begin{equation}
\begin{aligned}
    Con_{k} &= \dfrac{L_{intra}(I_{k})}{L_{inter}(I_{k})}\\ &=\dfrac{\sum_{i = 1;i \neq k}^m\Vert I_{k} - I_{i}\Vert_{2}^2}{\sum_{j = 1;j \neq C}^N\sum_{i = 1; y_{i} = j}^m\Vert I_{k} - I_{i}\Vert_{2}^2} . \\
\end{aligned}
\end{equation}

In particular, the interference from class $j$ can be expressed by $L_{inter-j}(I_{k})$, its total distance to other images in the class $j$ group. $L_{inter-j}(I_{k})$ is defined as
\begin{equation} 
    L_{inter-j}(I_{k}) = \sum_{i = 1;y_{i} = j}^m\Vert I_{k} - I_{i}\Vert_{2}^2 . 
\end{equation}

Based on (4) and (7), the class-separated indicator $Con_{k-j}$ to analyze the feature concentration degree for image $I_{k}$ with other class $j$ is defined as follows
\begin{equation}
\begin{aligned}
    Con_{k-j} &= \dfrac{L_{intra}(I_{k})}{L_{inter-j}(I_{k})}\\ &=\dfrac{\sum_{i = 1;i \neq k}^m\Vert I_{k} - I_{i}\Vert_{2}^2}{\sum_{i = 1;y_{i} = j}^m\Vert I_{k} - I_{i}\Vert_{2}^2} . \\
\end{aligned}
\end{equation}

In our assumption, samples with larger $Con_{k}$ are more likely to be mislabeled as noisy samples. For instance, When $Con_{k}$ of one sample is larger, it implies that its $L_{intra}(I_{k})$ is larger and $L_{inter}(I_{k})$ is smaller, thus indicating that the feature of the sample is farther from the feature group of its own class label and the sample is more similar to other categories.

\subsubsection{Validation Experiments}
To validate our assumption, we make a statistical analysis on the $Con_{k}$ of the "worst" set of CIFAR-10N, in which the noise ratio is estimated around 40\%. Specifically, we perform statistical analysis on the frequency distribution for $Con_{k}$ of each category in five intervals. Then, we mark the noisy samples among all samples as red ones and thus obtain the noise ratios in each interval. The results of the analysis are shown in Figure 8. Specifically, the histogram of the frequency distribution of $Con_{k}$ is shown in Figure 8 (a) and the corresponding line graph of five noise ratios is shown in Figure 8 (b).

\begin{figure*}[!h]
  \centering
\subfloat[{The histogram of the frequency distribution of $Con_{k}$}]{\label{fig:a}\includegraphics[width=6.4in]{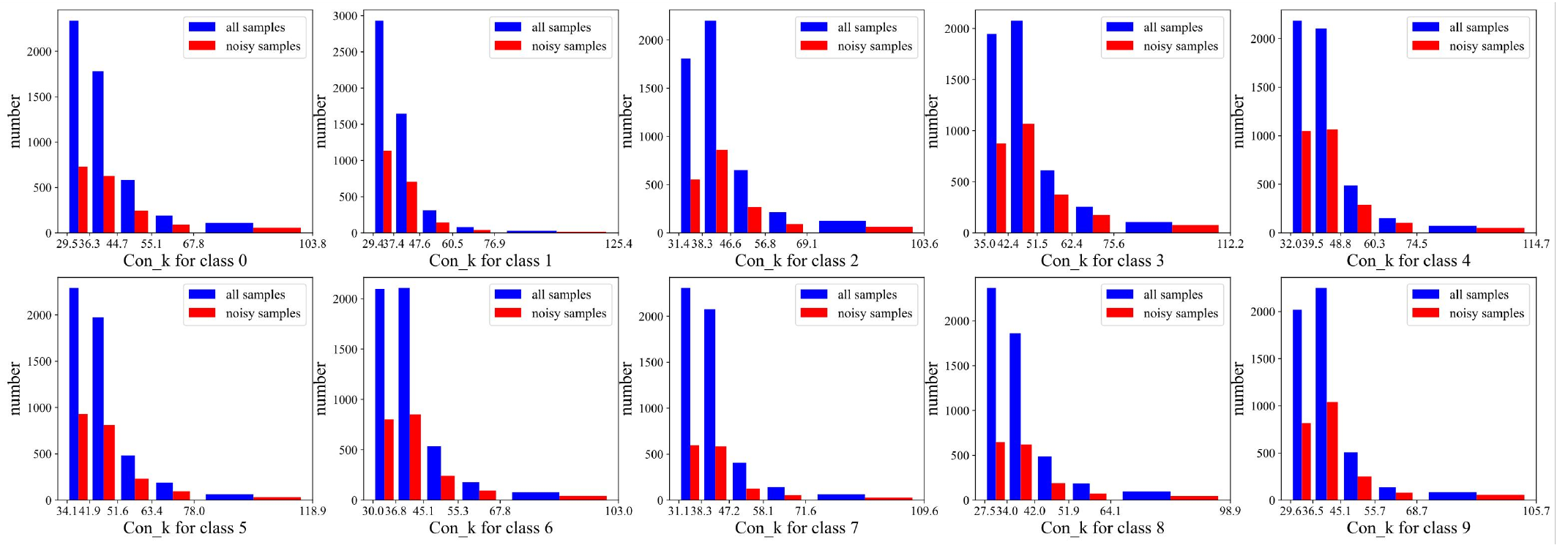}}\\
\subfloat[{The corresponding line graph of five noise ratios}]{\label{fig:b}\includegraphics[width=6.4in]{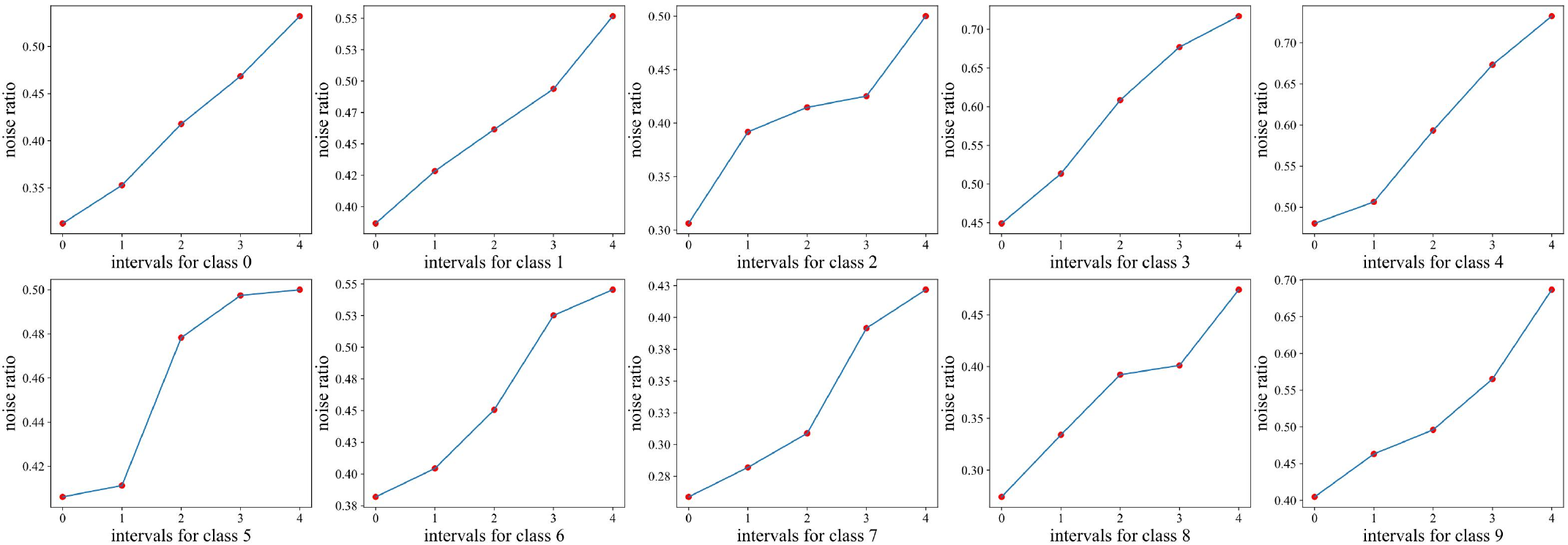}}\\	
	\caption{The results of the analysis on the "worst" set of CIFAR-10N}
  
\end{figure*}

We can see from the line graph that there is a growing trend with the increase of $Con_{k}$ value, but the noise ratio of the largest interval is not close to 1. This demonstrates that samples with a high value of $Con_{k}$ are more likely to be noisy samples but not necessarily noisy ones. Moreover, the variation of the growth slope in different categories is different. The noise ratio increases gently at the beginning and the end of the intervals but rapidly at the medium intervals for some categories. While for some categories, the situation is reversed. For other categories, the growth trend can be linear.

In order to observe the feature distribution in other synthetic noisy patterns, we make further experiments on the synthetic noisy CIFAR-10 dataset with IDN, the synthetic noisy CIFAR-10 dataset with symmetric noise and asymmetric noise in 40\% noise ratio. A similar statistical analysis is performed, and the results are shown in Figures 9,10,11.

It can be seen that the feature distribution of the real-world noise pattern perfectly confirms our assumption, while symmetric and asymmetric synthetic noise patterns don't. Since the IDN synthetic noise pattern is also designed based on the feature of instances, the growing trend of IDN noise pattern seems to be consistent with our assumption. However, we can notice that the distribution still doesn't exactly fit the real-world distribution for its extremely high upper bound and low lower bound. Thus, we can take this feature distribution indicator $Con_{k}$ as one of the reflections of the real-world noise pattern.

\subsubsection{Noisy Pattern Definition}
Based on these assumptions and discoveries, in this paper we define noise patterns by two factors, i.e., the noise transition matrix and the feature distribution. Therefore, we can obtain the real noise pattern by obtaining these two factors.

Firstly, the noise transition matrix $T_{s}$ for $D_{s}$ can be obtained by the clean labels $Y_{sc}$ and the noisy labels $Y_{sn}$ of $D_{s}$. By evaluating the transition matrix, we can acquire the noise ratio $\rho$ = $\{\rho_{1}, \rho_{2}, ..., \rho_{j}, ...\}$ in each category and the class label flipping probability $p_{j-1}$.

In addition, features $F_{s}$ can be extracted through the training process by the pre-trained ResNet34 model on $D_{s}$. Subsequently, inspired by \cite{zhu2019feature}, we propose an indicator to analyze the feature distribution. We think that it is the degree of concentration among an image, its class center and other class centers that impact the possibility of an image being mislabeled.

The feature distribution indicator $Con_{k}$ for each image $I_{k}$ in $D_{s}$ thus can be calculated based on $F_{s}$. Then we perform statistical analysis on all $Con_{k}$ in $D_{s}$. Separated by categories, for each class $j$, $j\in\{1, 2, 3, ..., N\}$, all $Con_{k}$ with class label $j$ are sorted from smallest to largest and divided into five intervals as shown in Figure 8. Note that the width of the interval is not uniform but gradually increasing.

\subsection{Noisy Data Generation with RGN Patterns}
 The expected overall noise ratio of the dataset is given as $\rho_{0}$. We can attain the total number of noisy samples $Num_{all}$ according to $\rho_{0}$ and $D_{c}$. The expected noise rate of each category in $D_{c}$ can be calculated according to the noise ratio $\rho$ = $\{\rho_{1}, \rho_{2}, ..., \rho_{N}\}$ of each category in $D_{s}$:
\begin{equation}
    R_{j} = \dfrac{Nc_{j}\times\rho_{j}}{\sum_{j=1}^NNc_{j}\times\rho_{j}} , 
\end{equation}
where $Nc_{j}$ is the number of samples with class label $j$ in $D_{c}$.
Then we obtain the number of noisy samples for each category:

\begin{equation} 
    Num_{j} = Num_{all} \times R_{j} . 
\end{equation}

 Afterward, combining with the clean labels $Y_{sc}$ and the noisy labels $Y_{sn}$ of $D_{s}$, the number of noisy samples in each feature interval for different categories can be obtained, denoted as $\{num_{j-i}\}$, where $j$ is the class label, $i\in\{1, 2, ..., 5\}$. Furthermore, the noise rate in each feature interval can be calculated by 
\begin{equation}
    r_{j-i} = \dfrac{num_{j-i}}{\sum_{i=1}^5num_{j-i}} . 
\end{equation}
 
 Given the clean set $D_{c}$, features $F_{c}$ also can be extracted through the training process by the pre-trained ResNet34 model on $D_{c}$. $Con_{k}$ for $D_{c}$ can be obtained in a similar way as $D_{s}$. Similarly, all $Con_{k}$ with class label $j$ are sorted from smallest to largest and divided into five intervals. Sample sets in five intervals are denoted as $\{S_{j-i}\}, i\in\{1, 2, ..., 5\}$.

Then the number of noisy samples in each feature interval for different categories can be obtained according to $\{Num_{j}\}$ and $\{r_{j-i}\}$:
\begin{equation} 
    Num_{j-i} = Num_{j} \times r_{j-i} . 
\end{equation}


Subsequently, $Num_{j-i}$ of samples are randomly selected from the $i_{th}$ interval in class $j$ as noisy samples $S_{n}$.

As for the selection of noisy labels for $S_{n}$, the clean labels of $S_{n}$ will be replaced based on (8). Specifically, if $I_{k}$ belongs to the noisy set $S_{n}$, for image $I_{k}$, $Con_{k-j}$ for each category will be calculated. The class label of the largest $Con_{k-j}$ will be converted as flipping probability $p_{j-2}$:

\begin{equation}
    p_{j-2} =\dfrac{Con_{k-j}}{\sum_{j = 1}^NCon_{k-j}} . \\
\end{equation}

 Taking both $p_{1}$ and $p_{2}$ into consideration, we calculate the final flipping probability as:

\begin{equation} 
    p_{j} = p_{j-1} \times \mu_{1} + p_{j-2} \times \mu_{2} , 
\end{equation}
where $\mu_{1}$ and $\mu_{2}$ are the balance hyper-parameters for the noise transition matrix and the feature distribution. We set $\mu_{1}$ as 0.1 and $\mu_{2}$  as 0.9 in our experiment. Ultimately, the noisy label of $I_{k}$ is selected as the class label $j$ of the largest $p_{j}$. Algorithm 1 shows the details of the noisy data generation process. Also, an overview of the generation process is shown in Figure 7.

\newcommand\mycommfont[1]{\footnotesize\ttfamily\textcolor{blue}{#1}}
\SetCommentSty{mycommfont}

\SetKwInput{KwInput}{Input}                
\SetKwInput{KwOutput}{Output}              

\begin{algorithm}[H]
\DontPrintSemicolon
  
  \KwInput{$F{c}=\{f_{c1},f_{c2},...,f_{cN}\}$, $f_{ci}$ is image feature of the clean set $D_{c}$, $F{s}=\{f_{s1},f_{s2},...,f_{sM}\}$, $f_{si}$ is image feature of the noisy subset $D_{s}$, noise transition matrix of subset $T_{s}$, clean labels $Y_{sc}$ for $D_{s}$, noisy labels $Y_{sn}$ for $D_{s}$, expected noise ratio $\rho_{0}$}
  \KwOutput{Noisy samples $X_{n}=\{x_{1},x_{2},...,x_{n}\}$ and corresponding noisy labels $Y_{n}=\{y_{1},y_{2},...,y_{n}\}$}
  Attain the total number of noisy samples $Num_{all}$ in $D_{c}$ based on $\rho_{0}$ and the number of samples in $D_{c}$
  
  Obtain the noise ratio of each category $\{\rho_{1}, \rho_{2}, ..., \rho_{N}\}$ and the
class label flipping probability $p_{j-1}$ from the noise transition matrix $T_{s}$. Determine the noise rate $R_{j}$, for each category in $D_{c}$ by (9)

  Obtain the number of noisy samples for each category 
 $Num_{j}$ based on $Num_{all}$ and $R_{j}$ by (10)
  
  Calculate the concentration degree of $F_{s}$ as $Con_{sk}$ based on (6), separate $Con_{sk}$ by categories as $Con_{sk(j)}$, sort $Con_{sk(j)}$ ascending, divide sorted $Con_{sk(j)}$ into five intervals, get the number of noisy samples in each feature interval with $Y_{sc}$ and $Y_{sn}$, as $num_{j-i}$, $i\in\{1, 2, ..., 5\}$, calculate the noise rate in each interval as $r_{j-i}$ by (11)
  
  In a similar way,  calculate the concentration degree of $F_{c}$ as $Con_{ck}$ based on (6). Sort $Con_{ck(j)}$ ascending; divide into five intervals. Sample sets in five intervals are denoted as $\{S_{j-i}\}, i\in\{1, 2, ..., 5\}$. Get the number of noisy samples in each interval as $Num_{j-i}$ by (12). Select randomly $Num_{j-i}$ samples from the $i_{th}$ interval, get noisy samples $X_{n}$

  Calculate $Con_{k-j}$ for $S_{n}$, sort $Con_{k-j}$ descending. The largest $Con_{k-j}$ is converted to $p_{j-2}$ by (13). The final flipping probability $p_j$ is calculated by (14). Choose the corresponding label of the largest $p_j$ as $Y_{n}$, as the noisy label of $X_{n}$
  
  \textbf{Return} $X_{n}$, $Y_{n}$

\caption{Noisy Data Generation}
\end{algorithm}

\begin{figure*}[!h]
  \centering
  \begin{minipage}[h]{1\linewidth}
    \centering
      \label{fig:subfig10}\includegraphics[width=6.4in]{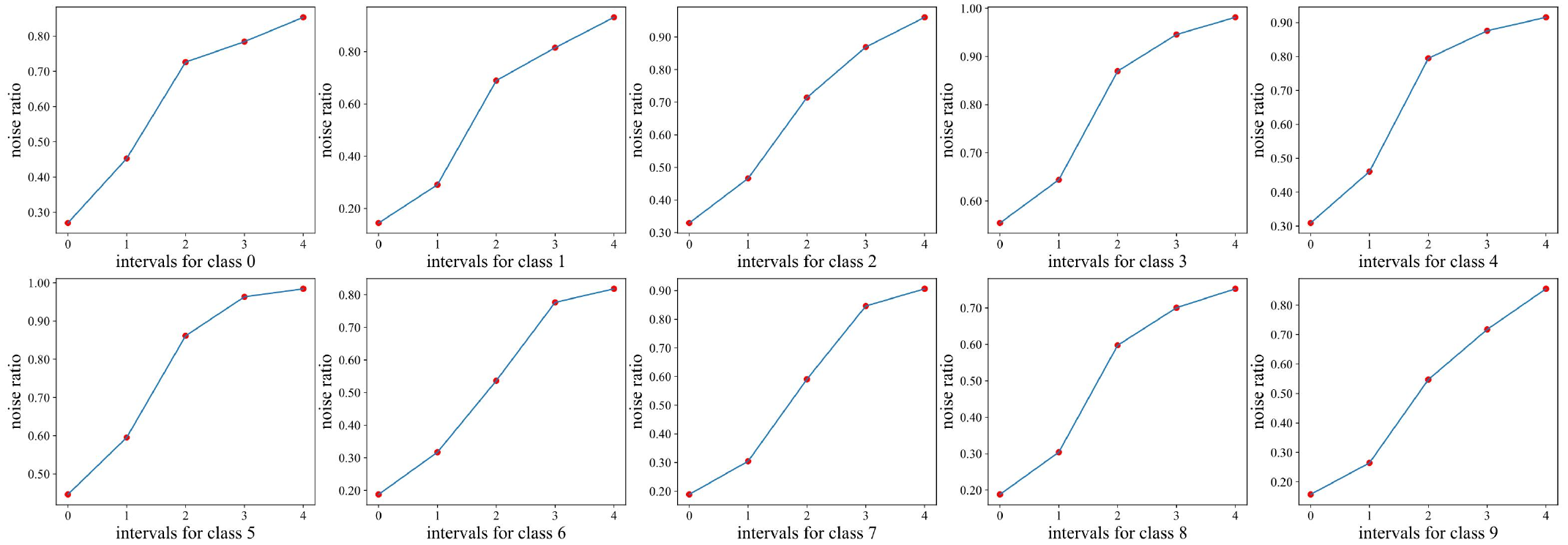}
      \caption{{The corresponding line graph of five noise ratios in the synthetic noisy CIFAR-10 dataset with IDN}}
  \end{minipage}
  \begin{minipage}[h]{1\linewidth}
    \centering
      \label{fig:subfig11}\includegraphics[width=6.4in]{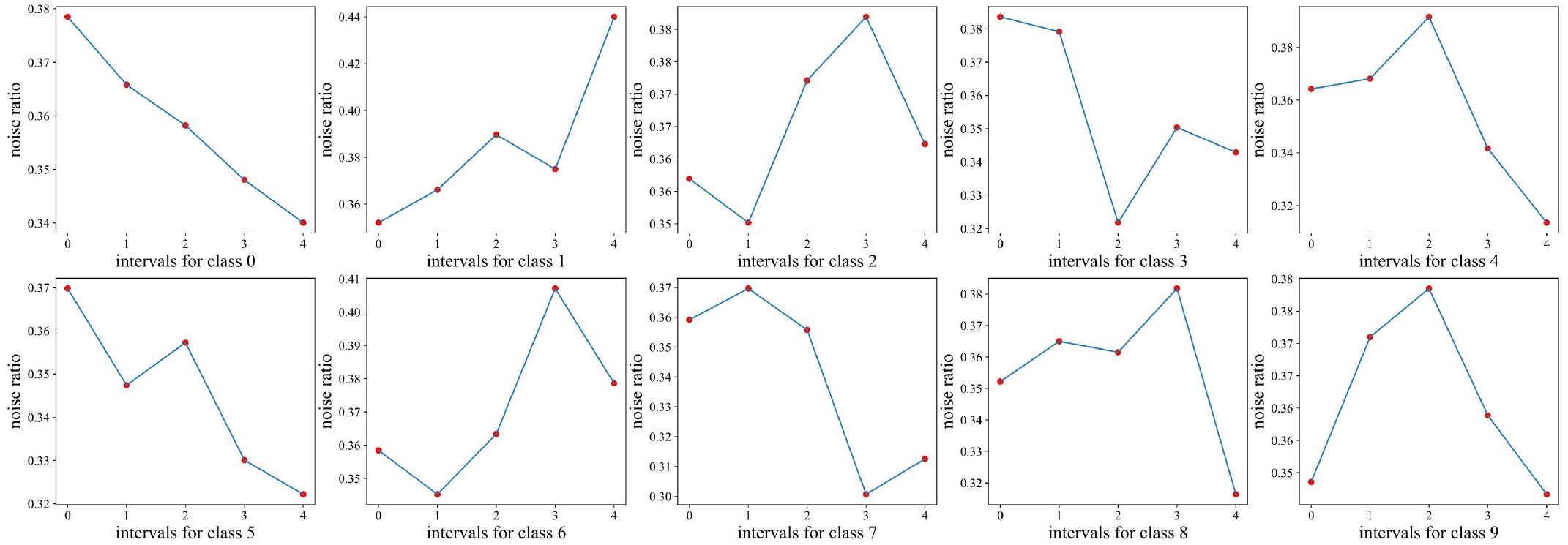}
      \caption{{The corresponding line graph of five noise ratios in the synthetic noisy CIFAR-10 dataset with 40\% symmetric noise}}
  \end{minipage}
  \begin{minipage}[h]{1\linewidth}
    \centering
      \label{fig:subfig12}\includegraphics[width=6.4in]{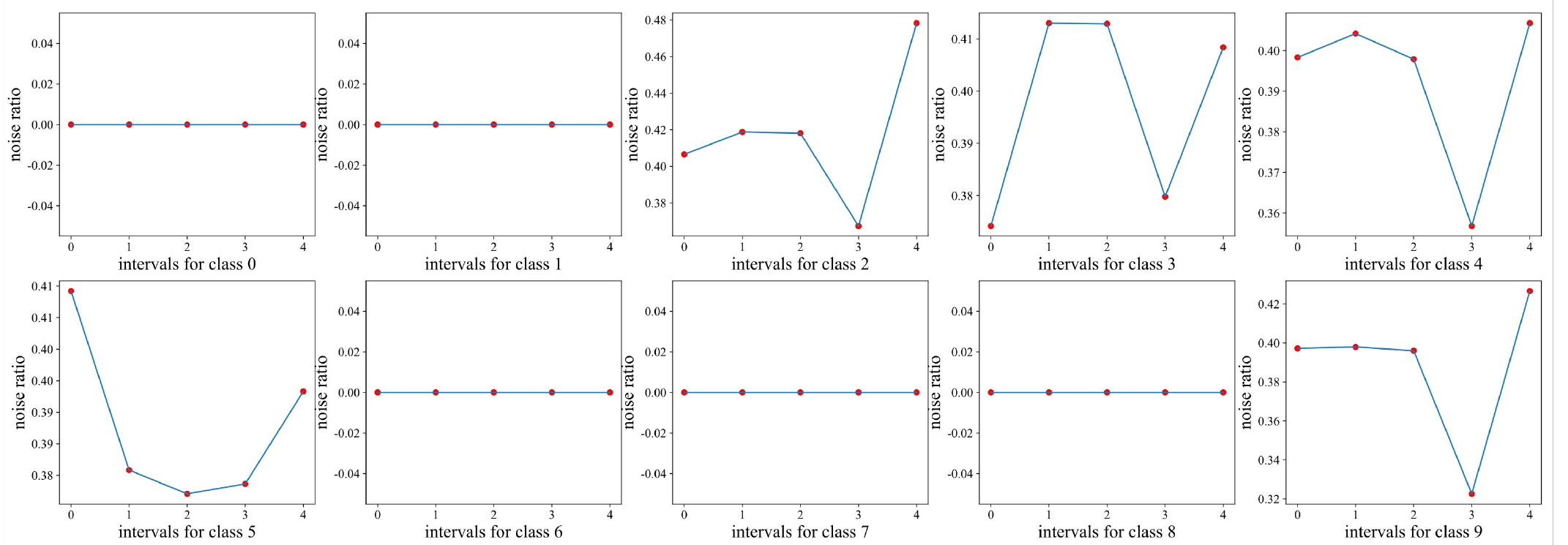}
      \caption{{The corresponding line graph of five noise ratios in the synthetic noisy CIFAR-10 dataset with 40\% symmetric noise}}
  \end{minipage}
  
\end{figure*}
\section{Experiments}
\subsection{Experimental Setup}
Nine representative methods of different categories, which are widely used for the image classification task in the presence of noisy labels, are selected for algorithm testing on the new benchmark we proposed, i.e., \emph{RoG} \cite{lee2019robust}, \emph{CDR} \cite{xia2020robust}, \emph{Mixup} \cite{zhang2017mixup}, \emph{SEAL} \cite{chen2021beyond}, \emph{Active passive loss} \cite{ma2020normalized}, \emph{JoCoR} \cite{wei2020combating}, \emph{Dividemix} \cite{li2020dividemix}, \emph{LongReMix} \cite{cordeiro2023longremix}, \emph{DISC} \cite{li2023disc}. In particular, LongReMix and DISC are state-of-the-art algorithms combating noisy labels. For fair comparisons, all methods use an 18-layer PreAct ResNet as the backbone and are trained with a GTX Tesla T4 GPU. Following past experimental settings, three noise ratios are adopted: 20\%, 50\%, and 80\%. 

The widely used evaluation metric for image classification is overall accuracy, which stands for the overall correcting performance in all samples regardless of the classes. Specifically, the overall accuracy is defined as the percentage of the correctly predicted samples in the total number of all samples, calculated by 
\begin{equation}
    acc = \dfrac{\sum_{i=1}^Cm_{i}}{\sum_{i=1}^Cn_{i}},
\end{equation}
where $m_{i}$ is the number of correctly predicted samples in class $i$ and $n_{i}$ is the total number of samples in class $i$. $C$ stands for the number of all classes.
\subsection{Experimental Results}
The experimental results on our proposed benchmark, CIFAR-10 with real-world guided noisy labels, are shown in Table 1. The overall accuracy of all testing methods can be seen as the baseline of our benchmark for further research in the field of combating noisy labels for classification tasks. In particular, both the best test accuracy across all training epochs and the average accuracy over the last ten epochs are reported in Table 1.

\begin{table}[h]
\caption{\textbf{Experiments on CIFAR-10 with RGN}}
\centering
\begin{tabular}{ccccc}
\toprule
&\multicolumn{2}{c}{\textbf{\underline{Result Summary}}}& \\
Method/Noise ratio&  & 20\% & 50\% & 80\% \\
\midrule
\multirow{2}{*}{RoG}&best&84.49&65.37&28.74 \\
&last \\
\midrule
\multirow{2}{*}{CDR}&best&87.39&77.18&31.08 \\
&last&84.94&71.24&25.17 \\
\midrule
\multirow{2}{*}{Mixup}&best&89.38&76.15&33.80 \\
&last&83.41&71.96&23.62 \\
\midrule
\multirow{2}{*}{SEAL}&best&90.30&82.39&26.49 \\
&last&90.05&81.86&16.82 \\
\midrule
\multirow{2}{*}{Active passive loss}&best&57.60&35.52&24.73 \\
&last&57.36&35.18&23.71 \\
\midrule
\multirow{2}{*}{JoCoR}&best&88.52&76.81&25.68 \\
&last&87.74&76.56&24.49 \\
\midrule
\multirow{2}{*}{DivideMix}&best&91.58&89.90&31.40 \\
&last&90.78&89.15&24.49 \\
\midrule
\multirow{2}{*}{LongReMix}&best&93.44&90.03&36.09 \\
&last&93.07&89.37&22.38 \\
\midrule
\multirow{2}{*}{DISC}&best&93.86&85.11&31.23 \\
&last&93.58&81.72&8.95 \\
\bottomrule
\end{tabular}
\end{table}

It is clear that \emph{LongReMix} obtains mostly the best performance over all noise ratios, steadily outperforming other algorithms. Specifically, it attains an overall accuracy of 93.44\% for 20\% noise ratio, 90.03\% for 50\% noise ratio, and 36.09\% for 80\% ratio. Following this, \emph{DISC} achieves an outstanding performance in 20\% noise ratio, slightly surpassing \emph{LongReMix} by 0.42\% and a comparable result in 50\% and 80\% ratio. Similarly, \emph{Dividemix}, also in the semi-supervised learning category, gets second place in the ranking of 50\% ratio and has great performance in other ratios. All three methods adopt the hybrid strategy of sample selection, semi-supervised learning, and other regularization skills, showing the effectiveness of the semi-supervised learning strategy. In particular, the remarkable result of \emph{LongReMix} indicates the necessity of precise selection of the clean subset. Moreover, \emph{SEAL}, belonging to the loss adjustment category, behaves greatly in three noise ratios. This may originate from its robust design for combating instance-dependent noisy labels. The normal sample selection method, \emph{JoCoR} shows an ordinary result over all ratios. The traditional kinds of methods like \emph{RoG}, \emph{CDR} and \emph{Mixup}, belonging to noise transition matrix estimation and robust regularization respectively, also show a plain performance. However, \emph{CDR} and \emph{Mixup}, the regularization kind of methods, achieve a comparable performance in 80\% noise ratio. This indicates that robust regularization skills may not work well in slight noise scenes but show great potential for the heavy noise scenario. For \emph{Active passive loss}, the robust loss function kind of method, it has the lowest and most terrible classification accuracy in the three ratios. This may be due to the change of the backbone. To find out the reason behind this phenomenon, we perform ablation experiments for \emph{Active passive loss} with the backbone network replaced by an 8-layer CNN and the noise type of RGN replaced by symmetric noise(settings used in the original paper of \emph{Active passive loss}). The experimental result is shown in Table 2. The result may illustrate that the single loss adjustment strategy alone is fragile and insufficient to combat complicated noisy labels. The design of this algorithm is only suitable for simple networks. What's more, it is worth noticing that all methods show poor performance under extreme noise ratio. This might be caused by the massive challenge of the combination of heavy noise and the confusing instance-based characteristic in \emph{RGN} noisy labels. Besides, we observe that under 80\% noise ratio, the best model appears in the warm-up period. Then the model performance gets worse and remains bad during training.

As we can see from the results, some methods that work well on symmetric label noise perform poorly on RGN, especially under extreme noise ratio, indicating that RGN is more challenging and close to reality than the current synthetic label noise patterns.

\begin{table}[H]
\caption{\textbf{Ablation study for the method of \emph{Active passive loss}}}
\centering
\begin{tabular}{cccc}
\toprule
Noise ratio/Module & 8-layer CNN & Symmetric& Accuracy \\
\midrule
\multirow{3}{*}{20\%}&\Checkmark& &86.67 \\
& &\Checkmark &59.66 \\
& & &57.60 \\
\midrule
\multirow{3}{*}{50\%}&\Checkmark& &75.30 \\
& &\Checkmark &38.90 \\
& & &35.52 \\
\midrule
\multirow{3}{*}{80\%}&\Checkmark& &27.29 \\
& &\Checkmark &25.40 \\
& & &24.73 \\
\bottomrule
\end{tabular}
\end{table}


\section{Conclusion}
The existence of noisy labels will have a greatly negative impact on the model generalization performance, and it is crucial to combat noisy labels for computer vision tasks, especially for classification tasks. To help better understand the evolution of research combating noisy labels, we first make a comprehensive review of different deep learning approaches for noise label combating in the image classification task, including four categories: noise transition matrix estimation, robust regularization, sample selection, semi-supervised learning based methods. In addition, we also review different noise patterns that have been proposed to design robust algorithms and categorize them into three taxonomies: instance-independent label noise, instance-dependent label noise, and human real label noise. Furthermore, we suggest two indicators, i.e., noise transition matrix and feature concentration distribution, to represent the
inner pattern of real-world label noise, as we called \emph{RGN}, and propose an algorithm to generate a synthetic label noise dataset guided by real-world data. We test the algorithm on the well-known real-world dataset CIFAR-10N to form a new synthetic benchmark guided by real-world data. We evaluate the typical methods of each category and the state-of-the-art methods in the task of image classification with
label noise on the benchmark we proposed for further research. The experimental results show that under the severe real-world noise pattern, most methods lose their effectiveness in maintaining the model robustness.

\section{Acknowledgement}
This work was supported by the National Key R\&D Program of China (2021ZD0109800), and by the High-performance Computing Platform of BUPT.

\bibliographystyle{IEEEtran}
\bibliography{IEEEabrv,IEEEexample}

\begin{thebibliography}{10}
\providecommand{\url}[1]{#1}
\csname url@samestyle\endcsname
\providecommand{\newblock}{\relax}
\providecommand{\bibinfo}[2]{#2}
\providecommand{\BIBentrySTDinterwordspacing}{\spaceskip=0pt\relax}
\providecommand{\BIBentryALTinterwordstretchfactor}{4}
\providecommand{\BIBentryALTinterwordspacing}{\spaceskip=\fontdimen2\font plus
\BIBentryALTinterwordstretchfactor\fontdimen3\font minus \fontdimen4\font\relax}
\providecommand{\BIBforeignlanguage}[2]{{%
\expandafter\ifx\csname l@#1\endcsname\relax
\typeout{** WARNING: IEEEtran.bst: No hyphenation pattern has been}%
\typeout{** loaded for the language `#1'. Using the pattern for}%
\typeout{** the default language instead.}%
\else
\language=\csname l@#1\endcsname
\fi
#2}}
\providecommand{\BIBdecl}{\relax}
\BIBdecl

\bibitem{krizhevsky2012imagenet}
A.~Krizhevsky, I.~Sutskever, and G.~E. Hinton, ``Imagenet classification with deep convolutional neural networks,'' \emph{NeurIPS}, vol.~25, pp. 1097--1105, 2012.

\bibitem{yu2018learning}
X.~Yu, T.~Liu, M.~Gong, and D.~Tao, ``Learning with biased complementary labels,'' in \emph{Proceedings of the European conference on computer vision (ECCV)}, 2018, pp. 68--83.

\bibitem{xie2019improving}
X.~Xie, J.~Mao, Y.~Liu, M.~de~Rijke, Q.~Ai, Y.~Huang, M.~Zhang, and S.~Ma, ``Improving web image search with contextual information,'' in \emph{Proceedings of the 28th ACM International Conference on Information and Knowledge Management}, 2019, pp. 1683--1692.

\bibitem{lloyd2004observer}
R.~V. Lloyd, L.~A. Erickson, M.~B. Casey, K.~Y. Lam, C.~M. Lohse, S.~L. Asa, J.~K. Chan, R.~A. DeLellis, H.~R. Harach, K.~Kakudo \emph{et~al.}, ``Observer variation in the diagnosis of follicular variant of papillary thyroid carcinoma,'' \emph{The American journal of surgical pathology}, vol.~28, no.~10, pp. 1336--1340, 2004.

\bibitem{zhang2021understanding}
C.~Zhang, S.~Bengio, M.~Hardt, B.~Recht, and O.~Vinyals, ``Understanding deep learning (still) requires rethinking generalization,'' \emph{Communications of the ACM}, vol.~64, no.~3, pp. 107--115, 2021.

\bibitem{frenay2013classification}
B.~Fr{\'e}nay and M.~Verleysen, ``Classification in the presence of label noise: a survey,'' \emph{IEEE transactions on neural networks and learning systems}, vol.~25, no.~5, pp. 845--869, 2013.

\bibitem{cordeiro2020survey}
F.~R. Cordeiro and G.~Carneiro, ``A survey on deep learning with noisy labels: How to train your model when you cannot trust on the annotations?'' in \emph{2020 33rd SIBGRAPI conference on graphics, patterns and images (SIBGRAPI)}.\hskip 1em plus 0.5em minus 0.4em\relax IEEE, 2020, pp. 9--16.

\bibitem{karimi2020deep}
D.~Karimi, H.~Dou, S.~K. Warfield, and A.~Gholipour, ``Deep learning with noisy labels: Exploring techniques and remedies in medical image analysis,'' \emph{Medical image analysis}, vol.~65, p. 101759, 2020.

\bibitem{song2022learning}
H.~Song, M.~Kim, D.~Park, Y.~Shin, and J.-G. Lee, ``Learning from noisy labels with deep neural networks: A survey,'' \emph{IEEE Transactions on Neural Networks and Learning Systems}, 2022.

\bibitem{kim2019nlnl}
Y.~Kim, J.~Yim, J.~Yun, and J.~Kim, ``Nlnl: Negative learning for noisy labels,'' in \emph{Proceedings of the IEEE/CVF international conference on computer vision}, 2019, pp. 101--110.

\bibitem{patrini2017making}
G.~Patrini, A.~Rozza, A.~Krishna~Menon, R.~Nock, and L.~Qu, ``Making deep neural networks robust to label noise: A loss correction approach,'' in \emph{Proceedings of the IEEE Conference on Computer Vision and Pattern Recognition}, 2017, pp. 1944--1952.

\bibitem{zhang2018generalized}
Z.~Zhang and M.~R. Sabuncu, ``Generalized cross entropy loss for training deep neural networks with noisy labels,'' in \emph{32nd Conference on Neural Information Processing Systems (NeurIPS)}, 2018.

\bibitem{xia2020part}
X.~Xia, T.~Liu, B.~Han, N.~Wang, M.~Gong, H.~Liu, G.~Niu, D.~Tao, and M.~Sugiyama, ``Part-dependent label noise: Towards instance-dependent label noise,'' \emph{Advances in Neural Information Processing Systems}, vol.~33, pp. 7597--7610, 2020.

\bibitem{zhang2021learning}
Y.~Zhang, S.~Zheng, P.~Wu, M.~Goswami, and C.~Chen, ``Learning with feature-dependent label noise: A progressive approach,'' \emph{arXiv preprint arXiv:2103.07756}, 2021.

\bibitem{chen2021beyond}
P.~Chen, J.~Ye, G.~Chen, J.~Zhao, and P.-A. Heng, ``Beyond class-conditional assumption: A primary attempt to combat instance-dependent label noise.'' in \emph{Proceedings of the AAAI Conference on Artificial Intelligence}, 2021.

\bibitem{wei2021learning}
J.~Wei, Z.~Zhu, H.~Cheng, T.~Liu, G.~Niu, and Y.~Liu, ``Learning with noisy labels revisited: A study using real-world human annotations,'' \emph{arXiv preprint arXiv:2110.12088}, 2021.

\bibitem{sukhbaatar2014training}
S.~Sukhbaatar, J.~Bruna, M.~Paluri, L.~Bourdev, and R.~Fergus, ``Training convolutional networks with noisy labels,'' \emph{arXiv preprint arXiv:1406.2080}, 2014.

\bibitem{srivastava2014dropout}
N.~Srivastava, G.~Hinton, A.~Krizhevsky, I.~Sutskever, and R.~Salakhutdinov, ``Dropout: a simple way to prevent neural networks from overfitting,'' \emph{The journal of machine learning research}, vol.~15, no.~1, pp. 1929--1958, 2014.

\bibitem{goldberger2016training}
J.~Goldberger and E.~Ben-Reuven, ``Training deep neural-networks using a noise adaptation layer,'' in \emph{5th International Conference on Learning Representations}, 2017.

\bibitem{chen2015webly}
X.~Chen and A.~Gupta, ``Webly supervised learning of convolutional networks,'' in \emph{Proceedings of the IEEE international conference on computer vision}, 2015, pp. 1431--1439.

\bibitem{hendrycks2018using}
D.~Hendrycks, M.~Mazeika, D.~Wilson, and K.~Gimpel, ``Using trusted data to train deep networks on labels corrupted by severe noise,'' \emph{Advances in neural information processing systems}, vol.~31, 2018.

\bibitem{bucarelli2023leveraging}
M.~S. Bucarelli, L.~Cassano, F.~Siciliano, A.~Mantrach, and F.~Silvestri, ``Leveraging inter-rater agreement for classification in the presence of noisy labels,'' in \emph{Proceedings of the IEEE/CVF Conference on Computer Vision and Pattern Recognition}, 2023, pp. 3439--3448.

\bibitem{xiao2015learning}
T.~Xiao, T.~Xia, Y.~Yang, C.~Huang, and X.~Wang, ``Learning from massive noisy labeled data for image classification,'' in \emph{Proceedings of the IEEE conference on computer vision and pattern recognition}, 2015, pp. 2691--2699.

\bibitem{han2018masking}
B.~Han, J.~Yao, G.~Niu, M.~Zhou, I.~Tsang, Y.~Zhang, and M.~Sugiyama, ``Masking: A new perspective of noisy supervision,'' \emph{Advances in neural information processing systems}, vol.~31, 2018.

\bibitem{yao2018deep}
J.~Yao, J.~Wang, I.~W. Tsang, Y.~Zhang, J.~Sun, C.~Zhang, and R.~Zhang, ``Deep learning from noisy image labels with quality embedding,'' \emph{IEEE Transactions on Image Processing}, vol.~28, no.~4, pp. 1909--1922, 2018.

\bibitem{lee2019robust}
K.~Lee, S.~Yun, K.~Lee, H.~Lee, B.~Li, and J.~Shin, ``Robust inference via generative classifiers for handling noisy labels,'' in \emph{International conference on machine learning}.\hskip 1em plus 0.5em minus 0.4em\relax PMLR, 2019, pp. 3763--3772.

\bibitem{collier2021correlated}
M.~Collier, B.~Mustafa, E.~Kokiopoulou, R.~Jenatton, and J.~Berent, ``Correlated input-dependent label noise in large-scale image classification,'' in \emph{Proceedings of the IEEE/CVF conference on computer vision and pattern recognition}, 2021, pp. 1551--1560.

\bibitem{xia2022extended}
X.~Xia, B.~Han, N.~Wang, J.~Deng, J.~Li, Y.~Mao, and T.~Liu, ``Extended $ t $ t: Learning with mixed closed-set and open-set noisy labels,'' \emph{IEEE Transactions on Pattern Analysis and Machine Intelligence}, vol.~45, no.~3, pp. 3047--3058, 2022.

\bibitem{han2019deep}
J.~Han, P.~Luo, and X.~Wang, ``Deep self-learning from noisy labels,'' in \emph{Proceedings of the IEEE/CVF international conference on computer vision}, 2019, pp. 5138--5147.

\bibitem{Tanno2020Learning}
R.~Tanno, A.~Saeedi, S.~Sankaranarayanan, D.~C. AleXaNder, and N.~Silberman, ``Learning from noisy labels by regularized estimation of annotator confusion,'' in \emph{2019 IEEE/CVF Conference on Computer Vision and Pattern Recognition (CVPR)}, 2020.

\bibitem{2020Early}
S.~Liu, J.~Niles-Weed, N.~Razavian, and C.~Fernandez-Granda, ``Early-learning regularization prevents memorization of noisy labels,'' 2020.

\bibitem{li2021improved}
D.~Li and H.~Zhang, ``Improved regularization and robustness for fine-tuning in neural networks,'' \emph{Advances in Neural Information Processing Systems}, vol.~34, pp. 27\,249--27\,262, 2021.

\bibitem{jenni2018deep}
S.~Jenni and P.~Favaro, ``Deep bilevel learning,'' in \emph{Proceedings of the European conference on computer vision (ECCV)}, 2018, pp. 618--633.

\bibitem{hendrycks2019using}
D.~Hendrycks, K.~Lee, and M.~Mazeika, ``Using pre-training can improve model robustness and uncertainty,'' in \emph{International conference on machine learning}.\hskip 1em plus 0.5em minus 0.4em\relax PMLR, 2019, pp. 2712--2721.

\bibitem{menon2019can}
A.~K. Menon, A.~S. Rawat, S.~J. Reddi, and S.~Kumar, ``Can gradient clipping mitigate label noise?'' in \emph{International Conference on Learning Representations}, 2019.

\bibitem{xia2020robust}
X.~Xia, T.~Liu, B.~Han, C.~Gong, N.~Wang, Z.~Ge, and Y.~Chang, ``Robust early-learning: Hindering the memorization of noisy labels,'' in \emph{International conference on learning representations}, 2020.

\bibitem{wei2021open}
H.~Wei, L.~Tao, R.~Xie, and B.~An, ``Open-set label noise can improve robustness against inherent label noise,'' \emph{Advances in Neural Information Processing Systems}, vol.~34, pp. 7978--7992, 2021.

\bibitem{goodfellow2014explaining}
I.~J. Goodfellow, J.~Shlens, and C.~Szegedy, ``Explaining and harnessing adversarial examples,'' \emph{arXiv preprint arXiv:1412.6572}, 2014.

\bibitem{zhang2017mixup}
H.~Zhang, M.~Cisse, Y.~N. Dauphin, and D.~Lopez-Paz, ``mixup: Beyond empirical risk minimization,'' \emph{arXiv preprint arXiv:1710.09412}, 2017.

\bibitem{pereyra2017regularizing}
G.~Pereyra, G.~Tucker, J.~Chorowski, {\L}.~Kaiser, and G.~Hinton, ``Regularizing neural networks by penalizing confident output distributions,'' \emph{arXiv preprint arXiv:1701.06548}, 2017.

\bibitem{lukasik2020does}
M.~Lukasik, S.~Bhojanapalli, A.~Menon, and S.~Kumar, ``Does label smoothing mitigate label noise?'' in \emph{International Conference on Machine Learning}.\hskip 1em plus 0.5em minus 0.4em\relax PMLR, 2020, pp. 6448--6458.

\bibitem{zhang2021delving}
C.-B. Zhang, P.-T. Jiang, Q.~Hou, Y.~Wei, Q.~Han, Z.~Li, and M.-M. Cheng, ``Delving deep into label smoothing,'' \emph{IEEE Transactions on Image Processing}, vol.~30, pp. 5984--5996, 2021.

\bibitem{lienen2021label}
J.~Lienen and E.~H{\"u}llermeier, ``From label smoothing to label relaxation,'' in \emph{Proceedings of the AAAI conference on artificial intelligence}, vol.~35, no.~10, 2021, pp. 8583--8591.

\bibitem{ko2023gift}
J.~Ko, B.~Yi, and S.-Y. Yun, ``A gift from label smoothing: robust training with adaptive label smoothing via auxiliary classifier under label noise,'' in \emph{Proceedings of the AAAI Conference on Artificial Intelligence}, vol.~37, no.~7, 2023, pp. 8325--8333.

\bibitem{qu2021dat}
Y.~Qu, S.~Mo, and J.~Niu, ``Dat: Training deep networks robust to label-noise by matching the feature distributions,'' in \emph{Proceedings of the IEEE/CVF Conference on Computer Vision and Pattern Recognition}, 2021, pp. 6821--6829.

\bibitem{cheng2023mitigating}
H.~Cheng, Z.~Zhu, X.~Sun, and Y.~Liu, ``Mitigating memorization of noisy labels via regularization between representations,'' in \emph{International Conference on Learning Representations (ICLR)}, 2023.

\bibitem{jiang2018mentornet}
L.~Jiang, Z.~Zhou, T.~Leung, L.-J. Li, and L.~Fei-Fei, ``Mentornet: Learning data-driven curriculum for very deep neural networks on corrupted labels,'' in \emph{International Conference on Machine Learning}.\hskip 1em plus 0.5em minus 0.4em\relax PMLR, 2018, pp. 2304--2313.

\bibitem{han2018co}
B.~Han, Q.~Yao, X.~Yu, G.~Niu, M.~Xu, W.~Hu, I.~Tsang, and M.~Sugiyama, ``Co-teaching: Robust training of deep neural networks with extremely noisy labels,'' in \emph{Advances in neural information processing systems}, 2018, pp. 8527--8537.

\bibitem{chen2019understanding}
P.~Chen, B.~Liao, G.~Chen, and S.~Zhang, ``Understanding and utilizing deep neural networks trained with noisy labels,'' \emph{arXiv preprint arXiv:1905.05040}, 2019.

\bibitem{huang2019o2u}
J.~Huang, L.~Qu, R.~Jia, and B.~Zhao, ``O2u-net: A simple noisy label detection approach for deep neural networks,'' in \emph{Proceedings of the IEEE/CVF International Conference on Computer Vision}, 2019, pp. 3326--3334.

\bibitem{malach2017decoupling}
E.~Malach and S.~Shalev-Shwartz, ``Decoupling" when to update" from" how to update",'' \emph{Advances in neural information processing systems}, vol.~30, 2017.

\bibitem{yu2019does}
X.~Yu, B.~Han, J.~Yao, G.~Niu, I.~Tsang, and M.~Sugiyama, ``How does disagreement help generalization against label corruption?'' in \emph{International Conference on Machine Learning}.\hskip 1em plus 0.5em minus 0.4em\relax PMLR, 2019, pp. 7164--7173.

\bibitem{wei2020combating}
H.~Wei, L.~Feng, X.~Chen, and B.~An, ``Combating noisy labels by agreement: A joint training method with co-regularization,'' in \emph{Proceedings of the IEEE/CVF conference on computer vision and pattern recognition}, 2020, pp. 13\,726--13\,735.

\bibitem{shen2019learning}
Y.~Shen and S.~Sanghavi, ``Learning with bad training data via iterative trimmed loss minimization,'' in \emph{International Conference on Machine Learning}.\hskip 1em plus 0.5em minus 0.4em\relax PMLR, 2019, pp. 5739--5748.

\bibitem{wang2018iterative}
Y.~Wang, W.~Liu, X.~Ma, J.~Bailey, H.~Zha, L.~Song, and S.-T. Xia, ``Iterative learning with open-set noisy labels,'' in \emph{Proceedings of the IEEE conference on computer vision and pattern recognition}, 2018, pp. 8688--8696.

\bibitem{song2021robust}
H.~Song, M.~Kim, D.~Park, Y.~Shin, and J.-G. Lee, ``Robust learning by self-transition for handling noisy labels,'' in \emph{Proceedings of the 27th ACM SIGKDD Conference on Knowledge Discovery \& Data Mining}, 2021, pp. 1490--1500.

\bibitem{wu2020topological}
P.~Wu, S.~Zheng, M.~Goswami, D.~Metaxas, and C.~Chen, ``A topological filter for learning with label noise,'' \emph{Advances in neural information processing systems}, vol.~33, pp. 21\,382--21\,393, 2020.

\bibitem{wu2021ngc}
Z.-F. Wu, T.~Wei, J.~Jiang, C.~Mao, M.~Tang, and Y.-F. Li, ``Ngc: A unified framework for learning with open-world noisy data,'' \emph{arXiv preprint arXiv:2108.11035}, 2021.

\bibitem{li2020dividemix}
J.~Li, R.~Socher, and S.~C. Hoi, ``Dividemix: Learning with noisy labels as semi-supervised learning,'' \emph{arXiv preprint arXiv:2002.07394}, 2020.

\bibitem{berthelot2019mixmatch}
D.~Berthelot, N.~Carlini, I.~Goodfellow, N.~Papernot, A.~Oliver, and C.~A. Raffel, ``Mixmatch: A holistic approach to semi-supervised learning,'' \emph{Advances in neural information processing systems}, vol.~32, 2019.

\bibitem{cordeiro2023longremix}
F.~R. Cordeiro, R.~Sachdeva, V.~Belagiannis, I.~Reid, and G.~Carneiro, ``Longremix: Robust learning with high confidence samples in a noisy label environment,'' \emph{Pattern Recognition}, vol. 133, p. 109013, 2023.

\bibitem{2021Augmentation}
K.~Nishi, Y.~Ding, A.~Rich, and T.~Hllerer, ``Augmentation strategies for learning with noisy labels,'' in \emph{Proceedings of the IEEE/CVF Conference on Computer Vision and Pattern Recognition (CVPR)}, 2021.

\bibitem{2022contrast}
E.~Zheltonozhskii, C.~Baskin, A.~Mendelson, A.~M. Bronstein, and O.~Litany, ``Contrast to divide: Self-supervised pre-training for learning with noisy labels,'' in \emph{Proceedings of the IEEE/CVF Winter Conference on Applications of Computer Vision}, 2022, pp. 1657--1667.

\bibitem{song2019selfie}
H.~Song, M.~Kim, and J.-G. Lee, ``Selfie: Refurbishing unclean samples for robust deep learning,'' in \emph{International Conference on Machine Learning}.\hskip 1em plus 0.5em minus 0.4em\relax PMLR, 2019, pp. 5907--5915.

\bibitem{nguyen2019self}
D.~T. Nguyen, C.~K. Mummadi, T.~P.~N. Ngo, T.~H.~P. Nguyen, L.~Beggel, and T.~Brox, ``Self: Learning to filter noisy labels with self-ensembling,'' \emph{arXiv preprint arXiv:1910.01842}, 2019.

\bibitem{zhou2020robust}
T.~Zhou, S.~Wang, and J.~Bilmes, ``Robust curriculum learning: From clean label detection to noisy label self-correction,'' in \emph{International Conference on Learning Representations}, 2020.

\bibitem{chen2023two}
M.~Chen, H.~Cheng, Y.~Du, M.~Xu, W.~Jiang, and C.~Wang, ``Two wrongs don’t make a right: Combating confirmation bias in learning with label noise,'' in \emph{Proceedings of the AAAI Conference on Artificial Intelligence}, vol.~37, no.~12, 2023, pp. 14\,765--14\,773.

\bibitem{kim2021joint}
Y.~Kim, J.~Yun, H.~Shon, and J.~Kim, ``Joint negative and positive learning for noisy labels,'' in \emph{Proceedings of the IEEE/CVF Conference on Computer Vision and Pattern Recognition}, 2021, pp. 9442--9451.

\bibitem{silva2022noise}
A.~Silva, L.~Luo, S.~Karunasekera, and C.~Leckie, ``Noise-robust learning from multiple unsupervised sources of inferred labels,'' in \emph{Proceedings of the AAAI Conference on Artificial Intelligence}, vol.~36, no.~8, 2022, pp. 8315--8323.

\bibitem{bai2021understanding}
Y.~Bai, E.~Yang, B.~Han, Y.~Yang, J.~Li, Y.~Mao, G.~Niu, and T.~Liu, ``Understanding and improving early stopping for learning with noisy labels,'' \emph{Advances in Neural Information Processing Systems}, vol.~34, pp. 24\,392--24\,403, 2021.

\bibitem{wang2022scalable}
Y.~Wang, X.~Sun, and Y.~Fu, ``Scalable penalized regression for noise detection in learning with noisy labels,'' in \emph{Proceedings of the IEEE/CVF Conference on Computer Vision and Pattern Recognition}, 2022, pp. 346--355.

\bibitem{chen2023softmatch}
H.~Chen, R.~Tao, Y.~Fan, Y.~Wang, J.~Wang, B.~Schiele, X.~Xie, B.~Raj, and M.~Savvides, ``Softmatch: Addressing the quantity-quality trade-off in semi-supervised learning,'' \emph{arXiv preprint arXiv:2301.10921}, 2023.

\bibitem{chen2021robustness}
P.~Chen, J.~Ye, G.~Chen, J.~Zhao, and P.-A. Heng, ``Robustness of accuracy metric and its inspirations in learning with noisy labels,'' in \emph{Proceedings of the AAAI Conference on Artificial Intelligence}, vol.~35, no.~13, 2021, pp. 11\,451--11\,461.

\bibitem{gui2021towards}
X.-J. Gui, W.~Wang, and Z.-H. Tian, ``Towards understanding deep learning from noisy labels with small-loss criterion,'' \emph{arXiv preprint arXiv:2106.09291}, 2021.

\bibitem{kim2021fine}
T.~Kim, J.~Ko, J.~Choi, S.-Y. Yun \emph{et~al.}, ``Fine samples for learning with noisy labels,'' \emph{Advances in Neural Information Processing Systems}, vol.~34, pp. 24\,137--24\,149, 2021.

\bibitem{wang2021tackling}
Q.~Wang, B.~Han, T.~Liu, G.~Niu, J.~Yang, and C.~Gong, ``Tackling instance-dependent label noise via a universal probabilistic model,'' in \emph{Proceedings of the AAAI Conference on Artificial Intelligence}, vol.~35, no.~11, 2021, pp. 10\,183--10\,191.

\bibitem{li2023disc}
Y.~Li, H.~Han, S.~Shan, and X.~Chen, ``Disc: Learning from noisy labels via dynamic instance-specific selection and correction,'' in \emph{Proceedings of the IEEE/CVF Conference on Computer Vision and Pattern Recognition}, 2023, pp. 24\,070--24\,079.

\bibitem{feng2023ot}
C.~Feng, Y.~Ren, and X.~Xie, ``Ot-filter: An optimal transport filter for learning with noisy labels,'' in \emph{Proceedings of the IEEE/CVF Conference on Computer Vision and Pattern Recognition}, 2023, pp. 16\,164--16\,174.

\bibitem{xu2023usdnl}
Y.~Xu, X.~Niu, J.~Yang, S.~Drew, J.~Zhou, and R.~Chen, ``Usdnl: uncertainty-based single dropout in noisy label learning,'' in \emph{Proceedings of the AAAI Conference on Artificial Intelligence}, vol.~37, no.~9, 2023, pp. 10\,648--10\,656.

\bibitem{yao2021jo}
Y.~Yao, Z.~Sun, C.~Zhang, F.~Shen, Q.~Wu, J.~Zhang, and Z.~Tang, ``Jo-src: A contrastive approach for combating noisy labels,'' in \emph{Proceedings of the IEEE/CVF conference on computer vision and pattern recognition}, 2021, pp. 5192--5201.

\bibitem{ortego2021multi}
D.~Ortego, E.~Arazo, P.~Albert, N.~E. O'Connor, and K.~McGuinness, ``Multi-objective interpolation training for robustness to label noise,'' in \emph{Proceedings of the IEEE/CVF Conference on Computer Vision and Pattern Recognition}, 2021, pp. 6606--6615.

\bibitem{karim2022unicon}
N.~Karim, M.~N. Rizve, N.~Rahnavard, A.~Mian, and M.~Shah, ``Unicon: Combating label noise through uniform selection and contrastive learning,'' in \emph{Proceedings of the IEEE/CVF Conference on Computer Vision and Pattern Recognition}, 2022, pp. 9676--9686.

\bibitem{li2022selective}
S.~Li, X.~Xia, S.~Ge, and T.~Liu, ``Selective-supervised contrastive learning with noisy labels,'' in \emph{Proceedings of the IEEE/CVF Conference on Computer Vision and Pattern Recognition}, 2022, pp. 316--325.

\bibitem{khosla2020supervised}
P.~Khosla, P.~Teterwak, C.~Wang, A.~Sarna, Y.~Tian, P.~Isola, A.~Maschinot, C.~Liu, and D.~Krishnan, ``Supervised contrastive learning,'' \emph{Advances in neural information processing systems}, vol.~33, pp. 18\,661--18\,673, 2020.

\bibitem{huang2023twin}
Z.~Huang, J.~Zhang, and H.~Shan, ``Twin contrastive learning with noisy labels,'' in \emph{Proceedings of the IEEE/CVF Conference on Computer Vision and Pattern Recognition}, 2023, pp. 11\,661--11\,670.

\bibitem{jiang2020beyond}
L.~Jiang, D.~Huang, M.~Liu, and W.~Yang, ``Beyond synthetic noise: Deep learning on controlled noisy labels,'' in \emph{International conference on machine learning}.\hskip 1em plus 0.5em minus 0.4em\relax PMLR, 2020, pp. 4804--4815.

\bibitem{zhu2019feature}
C.~Zhu, H.~Dong, and S.~Zhang, ``Feature fusion for image retrieval with adaptive bitrate allocation and hard negative mining,'' \emph{IEEE Access}, vol.~7, pp. 161\,858--161\,870, 2019.

\bibitem{ma2020normalized}
X.~Ma, H.~Huang, Y.~Wang, S.~Romano, S.~Erfani, and J.~Bailey, ``Normalized loss functions for deep learning with noisy labels,'' in \emph{International conference on machine learning}.\hskip 1em plus 0.5em minus 0.4em\relax PMLR, 2020, pp. 6543--6553.

\end{thebibliography}

\end{document}